\documentclass[11pt,letterpaper]{article}
\usepackage{techreport_style}

\def\TRInlineSupplementarySections{1}

\hypersetup{
  pdftitle={Training Agents to Evolve with Their Harness: A Deployed Digital Avatar Agent System},
  pdfauthor={TaoLive AIGC LLM Team},
  pdfsubject={Harness-aware post-training for production live-streaming digital avatar agents},
  pdfkeywords={agentic reinforcement learning, Harness Evolution, live streaming, digital avatars, technical report}
}

\title{Training Agents to Evolve with Their Harness: TaoLive Digital Avatar Agent Technical Report}
\teamname{TaoLive AIGC LLM Team}
\projectleaders{Meiguang Jin, Yibo Hu}
\corecontributors{Yuhan Sun$^*$, Wenhao Lin$^*$, Yibo Hu, Junfeng Ma, Meiguang Jin}
\contributors{Weihang Pan, Jiaxin Zhao, Zulong Chen}
\equalcontrib{$^*$\,Equal contribution}

\begin{document}
\maketitle

% Abstract
\begin{abstract}
Live e-commerce digital avatar streamers must answer product questions, interact with viewers, and execute marketing strategies in real time, requiring low latency, frequent strategy updates, and accurate yet effective replies. To meet the need for frequent updates, a common practice is to introduce an evolvable Harness, where Skills, Hooks, prompts, and tools can be modified independently of model weights. Ideally, this allows the system to iterate without retraining the model for every strategy change. However, this introduces a dilemma. While large models possess the generalization required to adapt to dynamic Harnesses zero-shot, their high latency prohibits real-time interaction. Conversely, low-latency compact models, typically trained on a fixed Harness, overfit to specific configurations and fail to evolve with Harness updates without continuous retraining.
We propose Harness-Aware Training (HAT) to train a compact model that can evolve with the Harness. The core idea is to expose the model to diverse Harness configurations during training so that it learns to understand the current Harness rather than memorize a fixed version. HAT introduces Harness-State Augmentation (HSA), which applies task-preserving transformations to Skill identifiers, Skill content, tool schemas, prompt structure, and Hook functions. Based on HSA, the training is organized into three stages: (1) HSA-SFT: the compact model learns from high-quality trajectories generated by strong models in diverse environments, directly improving its reasoning and tool-calling abilities. (2) General OPD: the model learns from the base model on general data via On-Policy Distillation to recover the generalizability damaged by SFT. (3) HSA-RL: reinforcement learning is applied in diverse augmented environments to further improve the model's understanding of changing Harnesses and its tool-calling skills, enhancing overall robustness.
Across four complementary evaluation sets, the HAT-trained model reaches an average score of 94.8 on Live-Stream QA (base 80.3, strongest general LLM 93.0) and 94.6 on Harness-Variant QA (base 75.4). Crucially, while traditional Fixed-Harness SFT causes a significant 7.7-point drop from the base model on IFEval, our method successfully avoids this degradation, achieving a strong score of 83.5. Furthermore, deployed on a single NVIDIA H20 GPU (with optimization enabled), it achieves a P50 latency of 3.4\,s and P95 of 8.1\,s, fully satisfying real-time deployment constraints. The system is also deployed in Taobao Live's production digital-avatar service. In an online A/B test, the Harness treatment recorded a UV-normalized uplift of 0.91\% in item-page views relative to the ReAct control.

\end{abstract}

\FloatBarrier
% 1. Introduction

\section{Introduction}

% P1: Real task and production requirements
Digital avatars are increasingly deployed across diverse interactive applications, spanning conversational assistants, customer service, education, and marketing~\citep{cui2023virtual, allalcherif2024intelligent, yang2025embodied}. Among these, live-streaming e-commerce represents a uniquely demanding scenario where AI-powered streamers must interact with viewers, answer product queries, and execute marketing strategies under strict real-time latency constraints~\citep{liu2025ai, sun2025livethinking, yu-etal-2026-taotype}. Since our live-streaming digital avatar’s replies are broadcast publicly, the underlying agent must simultaneously satisfy three stringent requirements: (1) ultra-low latency, essential for maintaining viewer engagement. (2) high adaptability to frequent shifts in campaign rules, compliance and merchant preferences. (3) accuracy and effectiveness, ensuring factually grounded responses that successfully address viewer intent without hallucination.
% Digital avatars are increasingly used in conversational assistants, customer service, education, and marketing~\citep{cui2023virtual, allalcherif2024intelligent, yang2025embodied}. Live-streaming e-commerce places additional real-time requirements on these systems: AI streamers must interact with viewers, answer product questions, and execute marketing strategies with low latency~\citep{liu2025ai, sun2025livethinking}. Because their responses are broadcast publicly, the underlying agent must operate with low end-to-end latency, adapt to frequent changes in campaign rules, compliance requirements, and merchant preferences, and produce factually grounded responses that address viewer intent without hallucination.

% P2: Why an evolvable Harness is needed
To achieve this agility without lengthy model retraining cycles, we adopt an evolvable Harness architecture—a modular paradigm related to recent advances in runtime optimization.~\citep{khattab2023dspy, hebbar2026sia, zhang2026selfharness, lin2026ahe, qu2026she, ning2025codeasharness, zhou2026selfevolvingcodingagents}. The Harness decouples the execution environment from the frozen policy model using four independently updatable modules: \textit{Skills} (reply rules and strategies), \textit{Hooks} (validation logic), a \textit{System Prompt Pipeline} (dynamic instructions), and a \textit{Tool Registry}. Through \textit{Harness Evolution}, a developer-reviewed diagnose–edit–evaluate loop applied to these modules, we can adjust business behaviors within hours.
% To update the agent's behavior without retraining it after every change, we use an evolvable Harness architecture related to recent work on runtime optimization~\citep{khattab2023dspy, hebbar2026sia, zhang2026selfharness, lin2026ahe, qu2026she, ning2025codeasharness}. The Harness separates the frozen policy model from four independently updatable runtime modules: \textit{Skills}, which define reply rules and strategies; \textit{Hooks}, which implement validation logic; a \textit{System Prompt Pipeline}, which assembles dynamic instructions; and a \textit{Tool Registry}. We call the developer-reviewed process for diagnosing failures, editing these modules, and evaluating the resulting configuration \textit{Harness Evolution}. This process allows us to revise business behavior within hours while keeping the model weights fixed.

% P3: 引出新的训练问题
 Harness Evolution also changes the training problem. The frozen policy model must operate across changing execution environments, including rewritten Skills and renamed tools, instead of relying on a single static configuration. Strong zero-shot models can interpret these changes without additional adaptation, but their latency is too high for our real-time setting. In our tests, DeepSeek-V4-flash~\citep{deepseekai2026deepseekv4} has a median end-to-end latency above 11 seconds. The latency requirement thus forces a compact policy model (e.g., Qwen3.6-35B-A3B~\citep{qwen3.6-35b-a3b}), which in turn must be domain-trained, because its zero-shot accuracy does not meet industrial standards. This is where the two mechanisms collide: training on a single configuration does improve domain task performance, yet it yields surface-form overfitting: the model memorizes specific skill names, tool names, and prompt templates rather than interpreting the actual instructions provided. Consequently, the model fails precisely when the Harness evolves. Experimentally, Fixed-Harness SFT degrades general instruction-following ability by 7.7 points on IFEval~\citep{zhou2023instruction}, a benchmark that measures a model's capacity to comply with explicit formatting and content constraints. Nor do existing alternatives address the issue: test-time adaptation~\citep{liang2024comprehensive, chen2026testtime} violates strict latency budgets, while continual learning~\citep{wang2024comprehensive} re-pays a retraining cost for every Harness edit, which is unsustainable at our update frequency.
% 参考：Harness Evolution also changes the training problem. The frozen policy model must operate across changing execution environments, including rewritten Skills and renamed tools, instead of relying on a single static configuration. Strong zero-shot models can interpret these changes without additional adaptation, but their latency is too high for our real-time setting. In our tests, DeepSeek-V4-flash has a median end-to-end latency above 11 seconds~\citep{deepseekai2026deepseekv4}. The latency requirement therefore favors a compact policy model such as Qwen3.6-35B-A3B~\citep{qwen3.6-35b-a3b}. However, its zero-shot accuracy does not meet our production requirements, making domain training necessary. Training a compact model under a single Harness improves domain performance but can lead to surface-form overfitting. The model may memorize Skill names, tool names, and prompt templates instead of interpreting the instructions in the active Harness. Its performance can therefore deteriorate after the Harness changes. In our experiments, Fixed-Harness SFT lowers IFE-P by 7.7 points on IFEval~\citep{zhou2023instruction}. Existing adaptation methods do not fit our operational constraints. Test-time adaptation adds inference-time computation that conflicts with the latency budget~\citep{liang2024comprehensive, chen2026testtime}, while continual learning requires additional training after Harness edits~\citep{wang2024comprehensive}, which is impractical at our update frequency.

% P4: Harness-Aware Training and its three stages
We address this issue with Harness-Aware Training (HAT), which makes the Harness state an explicit part of the training distribution rather than treating it as a static deployment-time constant. Instead of fitting the policy to trajectories collected under the single configuration that happens to be in production, we train over a distribution of Harness states, so that the model learns to condition on whatever Harness it is currently given and can evolve together with it. Concretely, we introduce Harness-State Augmentation (HSA), which applies task-preserving perturbations to Skill content, tool schemas, prompt structure, and interaction constraints, and instantiate it across a three-stage pipeline. (1)~\textit{SFT with HSA (HSA-SFT)}: the compact model learns directly from quality-filtered trajectories that a stronger teacher model generates across HSA-diversified Harness environments, directly sharpening its tool-calling and reasoning comprehension. (2)~\textit{General On-Policy Distillation (General OPD)}~\citep{agarwal2024onpolicy}: the model distills from the base model on general-domain data to recover the general capabilities degraded during domain-specific SFT~\citep{lu2025onpolicydistillation}. (3)~\textit{Agentic RL with HSA (HSA-RL)}: the model is further trained with reinforcement learning inside HSA-diversified environments within a production-informed live-room simulator, strengthening its comprehension of shifting Harness configurations and its tool-calling ability under those shifts, thereby improving overall robustness.
% 参考：We propose Harness-Aware Training (HAT), which treats the Harness state as a variable in the training distribution. Instead of training only on trajectories collected under the current production configuration, HAT trains the policy across a distribution of Harness states. This encourages the model to condition on the active configuration rather than memorize a fixed one. Harness-State Augmentation (HSA) generates task-preserving variants of Skill content, tool schemas, prompt structures, and interaction constraints. HAT then applies these variants in three stages. First, HSA-SFT trains the compact model on quality-filtered trajectories generated by a stronger teacher model under diverse Harness configurations. Second, General On-Policy Distillation (General OPD)~\citep{agarwal2024onpolicy} distills the base model's behavior on general-domain data to recover capabilities degraded by domain-specific SFT~\citep{lu2025onpolicydistillation}. Third, HSA-RL further optimizes the policy through reinforcement learning in a production-informed live-room simulator with augmented Harness configurations.

% P5: Offline results
Extensive evaluations across four complementary benchmarks totaling over 4{,}500 cases demonstrate the effectiveness of our approach. On our real-world Live-Stream QA benchmark, the HAT trained Qwen3.6-35B-A3B achieves a 94.8 average score, surpassing both the untuned base model (80.3) and strong zero-shot models. Crucially, it maintains robustness to evaluated Harness changes and avoids the significant IFEval score loss observed after Fixed-Harness SFT.
% 修改：We evaluate HAT on four benchmarks containing more than 4{,}500 cases. On Live-Stream QA, the HAT-trained Qwen3.6-35B-A3B scores 94.8, compared with 80.3 for the base model and 93.0 for the best evaluated zero-shot model. It scores 94.6 on Harness-Variant QA, compared with 75.4 for the base model. On IFEval, Fixed-Harness SFT lowers IFE-P from 81.5 to 73.8, whereas HAT reaches 83.5.
% P6: Deployment performance
Our Harness-based judge is calibrated against human annotations to ensure reliability. In controlled deployment tests on a single NVIDIA H20 GPU with MTP~\citep{deepseekv3, li2024eagle, chen2026dflash, cheng2026dspark}, the complete agent meets strict latency requirements, achieving a P50 wall-clock latency of 3.407s and a P95 of 8.114s.
We further deploy the system in Taobao Live's production digital-avatar service, where an online A/B test shows an increase in item-page views per participating user relative to ReAct.
% 修改：We calibrate the Harness-based Agent-as-a-Judge against human annotations. In a controlled complete-Agent deployment replay on a single NVIDIA H20 GPU with MTP enabled~\citep{deepseekv3, li2024eagle, chen2026dflash, cheng2026dspark}, HAT achieves P50 and P95 wall-clock latencies of 3.407s and 8.114s, respectively. The system is also deployed in Taobao Live's production digital-avatar service. In an online A/B test, the Harness treatment records a UV-normalized uplift of 0.9107% in item-page views relative to the ReAct control.

Our contributions are:
\begin{itemize}
    \item \textbf{An evolvable Harness architecture and its training problem.} We present an evolvable Harness architecture for live-streaming agents. The architecture separates the policy model from independently versioned Skills, Hooks, prompts, and tools, allowing business behavior to be updated without changing the model weights. We also formulate the training problem created by this changing execution environment (Section~\ref{sec:system}).
    % An evolvable Harness architecture and its training problem. We present an evolvable Harness architecture for live-streaming agents. The architecture separates the policy model from independently versioned Skills, Hooks, prompts, and tools, allowing business behavior to be updated without changing the model weights. We also formulate the training problem created by this changing execution environment.

    \item \textbf{Harness-Aware Training over changing Harness states.} We introduce Harness-Aware Training, which places Harness variation in the training distribution. HAT combines HSA-SFT, General OPD, and HSA-RL to train the policy to condition on the active Harness configuration instead of memorizing one version (Sections~\ref{sec:method}).
    % Harness-Aware Training over changing Harness states. We introduce Harness-Aware Training, which places Harness variation in the training distribution. HAT combines HSA-SFT, General OPD, and HSA-RL to train the policy to condition on the active Harness configuration instead of memorizing one version.

    \item \textbf{Evaluation of response quality, Harness variation, and general capability.} We construct four evaluation sets for live-streaming response quality, Harness variation, tool and prompt robustness, and general instruction following. We also calibrate the Harness-based Agent-as-a-Judge against human annotations and evaluate serving performance through a controlled complete-Agent replay (Section~\ref{sec:evaluation}).
    % Evaluation of response quality, Harness variation, and general capability. We construct four evaluation sets for live-streaming response quality, Harness variation, tool and prompt robustness, and general instruction following. We also calibrate the Harness-based Agent-as-a-Judge against human annotations and evaluate serving performance through a controlled complete-Agent replay.

    \item \textbf{From offline evaluation to production deployment.} HAT improves domain performance and robustness to the evaluated Harness changes while preserving general instruction-following ability. The deployed system meets the latency target, receives more preferences than ReAct in the human blind test, and records a UV-normalized uplift in item-page views in the Taobao Live A/B test (Sections~\ref{sec:experiments} and~\ref{sec:online_ab_taobao}).
    % From offline evaluation to production deployment. HAT improves domain performance and robustness to the evaluated Harness changes while preserving general instruction-following ability. The deployed system meets the latency target, receives more preferences than ReAct in the human blind test, and records a UV-normalized uplift in item-page views in the Taobao Live A/B test.
\end{itemize}

\FloatBarrier
% 2. System - original content + concise task/runtime details
\section{Digital-Avatar Harness Agent Architecture}
\label{sec:system}

The Harness Agent separates a slowly updated \emph{policy model} from a rapidly evolving \emph{Harness state}. This separation lets operators change business behavior without retraining. We first describe the application and runtime, then formalize the non-training process as \emph{Harness Evolution}.

\subsection{Live-Streaming Digital-Avatar Scenario}
\label{sec:application}

\begin{figure*}[t]
\centering
\includegraphics[width=\textwidth]{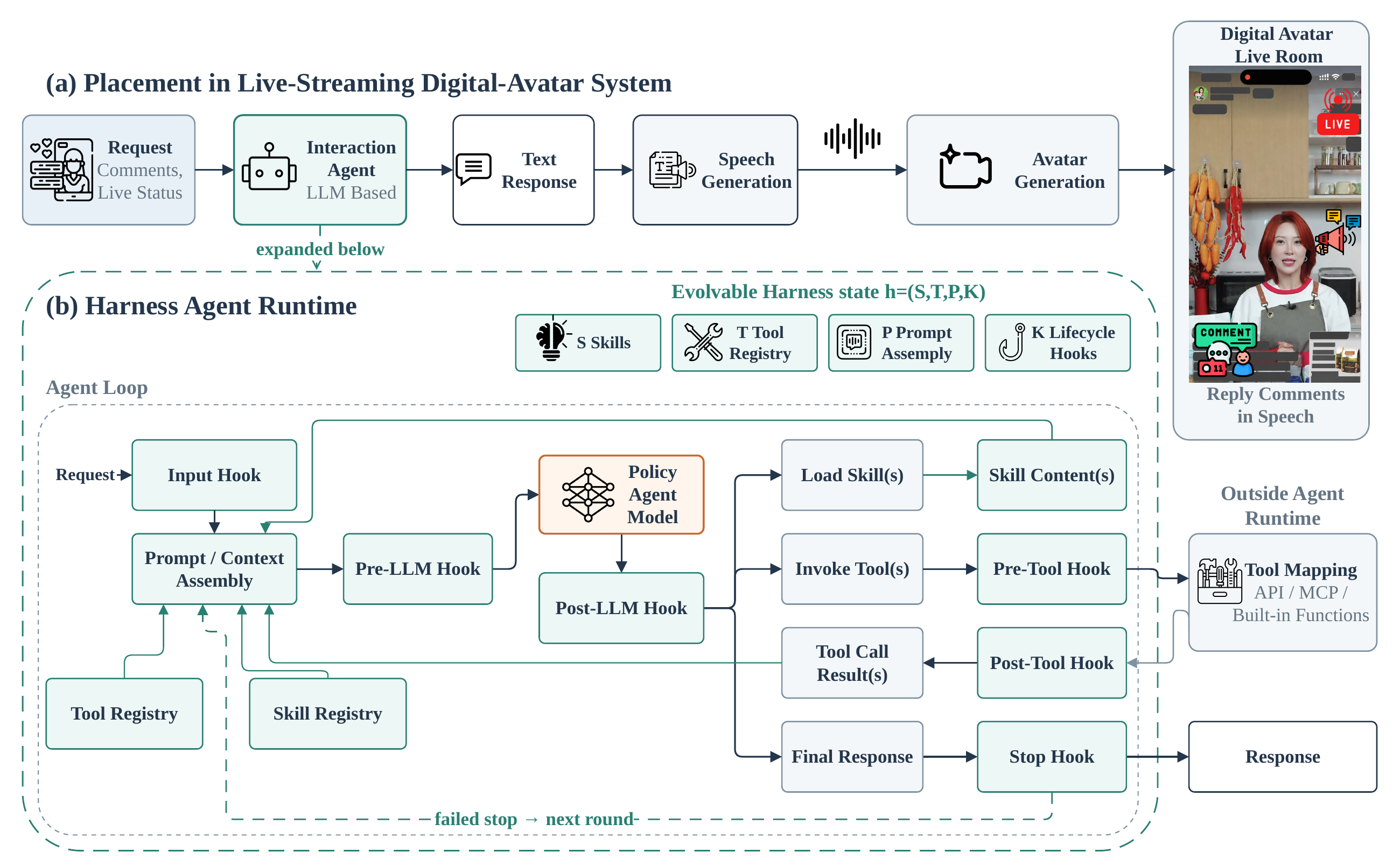}
\caption{Architecture of the live-streaming digital-avatar system and its Harness Agent runtime. (a) Viewer requests and live-room context are processed by the interaction agent. The resulting text response is converted to speech, rendered by the avatar generator, and broadcast to the live room. (b) The expanded runtime operates under an evolvable Harness state $h=(\mathcal{S},\mathcal{T},\mathcal{P},\mathcal{K})$, comprising active Skills, the tool registry, the assembled system prompt, and lifecycle Hooks, respectively. In each agent-loop round, Hooks mediate input processing, model inference, tool execution, and stopping, while the policy may load Skills, invoke externally mapped tools, or propose a final response. A successful stop check emits the response; a failed check returns control to context assembly for the next round.}
\label{fig:system_overview}
\end{figure*}

We present a digital-avatar host designed to answers product questions, engage in open-ended conversations, and invoke external tools during live-stream commerce. The avatar's responses are synthesized via a Text-to-Speech (TTS) module and synchronized with an audio-driven avatar generation model for broadcasting to the entire live room. To ensure contextual accuracy, each input request integrates the viewer's message with the anchor's dialogue history, product metadata, action link IDs, and the real-time status of the live room. Furthermore, the system necessitates ultra-low latency while seamlessly adapting to rapidly evolving business rules. As illustrated in Fig.~\ref{fig:system_overview} (a), our interactive agent operates as the core within a comprehensive digital avatar framework.It takes real-time audience comments and live-stream contextual data as inputs, processing them to formulate appropriate responses and executable actions. Subsequently, the backend server converts the agent's output into a multimodal format, utilizing TTS for audio synthesis and an avatar generation algorithm for video rendering. Finally, this synchronized audio-visual content is streamed to client devices, delivering a seamless live-streaming experience to the audience.

\subsection{Harness Agent Runtime}
\label{sec:runtime}

The Harness Agent runtime comprises four mechanisms. \emph{Skills} are dynamically loaded behavioral modules for Q\&A strategy, tool choosing, and reply style. The \emph{System Prompt Pipeline} assembles instructions from product specifications, retrieved FAQ, living-room status, global boundary and active skill descriptions. \emph{Hooks} validate inputs, parameters, and output formats, triggering correction or retry when needed. \emph{Tool Registry} connects the agent to external product, inventory, and commerce services. Because these modules are versioned independently of model weights, a new skill or tool can be deployed as a harness change rather than a retraining cycle. The overview of our harness agent architecture is shown in Fig.~\ref{fig:system_overview} (b).

\subsection{Harness Evolution without Weight Updates}
\label{sec:harness_evolution}

We introduce \emph{Harness Evolution} to denote a controlled process that improves the agent by changing skills, prompts, hooks, or tools while holding the model fixed. It is human-in-loop rather than fully autonomous: AI performs failure clustering, diagnosis, and proposed edits, a developer confirms the plan, runs evaluation, and decides whether to promote, revise, or stop. The complete operational protocol appears in Appendix~\ref{app:harness_evolution}.

\begin{definitionbox}{Harness Evolution Loop}
\centering
\textsc{Diagnose} $\rightarrow$ \textsc{Confirm} $\rightarrow$ \textsc{Edit Harness} $\rightarrow$ \textsc{Evaluate} $\rightarrow$ \textsc{Regression Check}
\end{definitionbox}

\begin{figure}[t]
\centering
\includegraphics[width=0.6\linewidth]{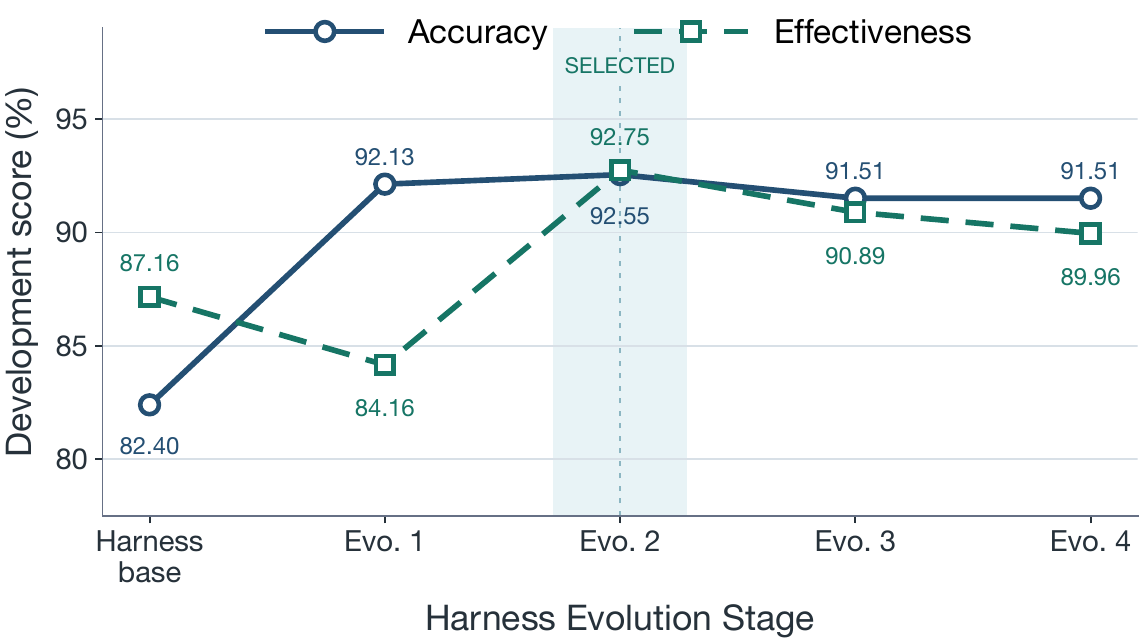}
\caption{Harness Evolution with a fixed DeepSeek-V4-flash on dev-set ($n=482$). Optimizing the agent without training model. Evolution~2 is the selected development checkpoint Harness, while later long-tail edits introduce regressions.}
\label{fig:harness_evolution_trace}
\end{figure}

Figure~\ref{fig:harness_evolution_trace} illustrates both the value of the \emph{Harness Evolution}. On the dev-set annotated by three annotators. Evolution~1 sharply improves Accuracy from 82.40 to 92.13 but reduces Effectiveness to 84.16. Evolution~2 reaches 92.55 Accuracy and 92.75 Effectiveness. Additional long-tail rules in Evolutions~3--4 regress one or both dimensions, so Evolution~2 is selected as the engineering early-stop checkpoint. Detailed module changes, category diagnostics, stopping logic, and evidence boundaries are reported in Appendix~\ref{app:harness_evolution}. The evaluation Judge is also implemented as an editable harness agent as detailed in Appendix~\ref{app:judge_evolution}.

% Downgrade shared file: \section->\subsection, \subsection->\subsubsection, \subsubsection->\paragraph
\begingroup
\def\TRdeferRuntimeInventoryTables{1}
\let\TRorigsection\section
\let\TRorigsubsection\subsection
\let\TRorigsubsubsection\subsubsection
\let\section\TRorigsubsection
\let\subsection\TRorigsubsubsection
\let\subsubsection\paragraph
\section{Task and Runtime Details}
\label{app:task_description}

This section expands the task interface and runtime inventory summarized in Section~\ref{sec:system}. Evaluation roles and scoring are specified separately in Appendix~\ref{app:evaluation_details}.

\subsection{Task Interface}
\label{app:task_def}

In single-comment mode, the runtime handles one viewer message independently. In multi-comments mode, it aggregates related messages before responding. An available internal traffic summary groups the main viewer intents as shown in Table~\ref{tab:scenario_dist}. 

\begin{table}[!b]
\centering
\begin{minipage}{0.78\linewidth}
\captionsetup{width=\linewidth}
\caption{Approximate intent mix in the available internal traffic summary.}
\label{tab:scenario_dist}
\small
\centering
\begin{tabular*}{\linewidth}{@{\extracolsep{\fill}}lc@{}}
\toprule
\tablehead{Scenario} & \tablehead{Proportion} \\
\midrule
Product Q\&A & $\sim$46\% \\
Casual chat and engagement & $\sim$19\% \\
Clarification follow-up & $\sim$16\% \\
After-sales handling & $\sim$7\% \\
Discount and promotion inquiry & $\sim$4\% \\
Presentation-order adjustment & $\sim$2\% \\
FAQ-based reply & $\sim$2\% \\
Silent refusal of irrelevant content & $<1\%$ \\
\bottomrule
\end{tabular*}
\end{minipage}
\end{table}

\subsection{Runtime Interfaces}
\label{app:tool_set}

The Harness exposes built-in interfaces for retrieval, commerce operations, and flow control. Additional marketing interfaces are discovered dynamically from an external service through the Model Context Protocol (MCP). Table~\ref{tab:runtime_interfaces} groups the release inventory by responsibility. 

\ifdefined\TRdeferRuntimeInventoryTables\else
\ifdefined\TRruntimeTable\else
\newenvironment{TRruntimeTable}{\begin{table*}[t]}{\end{table*}}
\fi
\begin{TRruntimeTable}
\centering
\caption{Working inventory of built-in and MCP-discovered runtime interfaces.}
\label{tab:runtime_interfaces}
\footnotesize
\setlength{\tabcolsep}{4pt}
\begin{tabularx}{\textwidth}{@{}p{2.35cm} >{\raggedright\arraybackslash}X >{\raggedright\arraybackslash}p{5.1cm}@{}}
\toprule
\tablehead{Responsibility} & \tablehead{Interfaces} & \tablehead{Role} \\
\midrule
Product retrieval &
\path{search_product_by_keyword}, \path{get_current_product_info}, \path{get_product_info_by_link_id}, \path{get_product_extra_info_by_link_id_and_keywords}, \path{search_preset_faq} &
Find the active or referenced product and retrieve catalog, detail-page, knowledge-base, or FAQ evidence. \\
Pricing and promotion &
\path{get_price_info_by_link_id}, \path{get_promotion_info} &
Retrieve SKU-level prices, entitlements, coupons, and room-level promotions. \\
Action and flow control &
\path{change_explain_order}, \path{load_skill}, \path{get_current_time}, \path{refuse_to_reply} &
Change presentation order, load behavioral instructions, resolve time-sensitive rules, or silently discard meaningless input. \\
MCP lookup &
\path{query_promotion}, \path{query_buyer_resource}, \path{query_fund_asset}, \path{query_coupon_detail} &
Retrieve campaign conditions and buyer-side resources from the external marketing service. \\
MCP calculation &
\path{calculate_optimal_promotion}, \path{calculate_promotion} &
Compute eligible promotional combinations and resulting prices. \\
\bottomrule
\end{tabularx}
\end{TRruntimeTable}

\fi

\subsection{Skill Routing}
\label{app:skill_system}

Each Skill is a versioned Markdown module with YAML metadata and behavioral directives. The runtime requires at least one Reply Skill before a substantive final response. Strategy Skills are optional and can be composed when an interaction requires a specialized tool chain. Table~\ref{tab:skill_inventory} records the release inventory. Module names and versions are reconciled against the same manifest before promotion.

\ifdefined\TRdeferRuntimeInventoryTables\else
\ifdefined\TRruntimeTable\else
\newenvironment{TRruntimeTable}{\begin{table*}[t]}{\end{table*}}
\fi
\begin{TRruntimeTable}
\centering
\caption{Working Skill inventory and routing responsibilities.}
\label{tab:skill_inventory}
\footnotesize
\setlength{\tabcolsep}{4pt}
\begin{tabularx}{\textwidth}{@{}>{\raggedright\arraybackslash}p{2.6cm} >{\raggedright\arraybackslash}X >{\raggedright\arraybackslash}p{3.25cm} >{\raggedright\arraybackslash}X@{}}
\toprule
\tablehead{Reply Skill} & \tablehead{Response responsibility} & \tablehead{Strategy Skill} & \tablehead{Tool-chain responsibility} \\
\midrule
\path{item_qa} & Product attributes, specifications, comparison, price, and recommendation. & \path{tool_strategy_benefit} & Entitlement, price, and promotion retrieval. \\
\path{chillchat} & Non-product conversation and engagement. & \path{tool_strategy_compare} & Multi-product comparison and retrieval. \\
\path{aftersale} & Returns, exchanges, complaints, and other after-sales requests. & \path{tool_strategy_multi} & Aggregation and orchestration for multiple comments. \\
\path{clarification} & Follow-up questions when available context is insufficient. & \path{tool_strategy_switch} & Presentation-order adjustment. \\
\path{faq_reply} & Responses grounded in configured FAQ entries. & \path{general_discount} & Room-level promotion retrieval and response composition. \\
\path{refusal} & Declines for inappropriate or unsupported requests. & & \\
\path{change_order} & Confirmation of presentation-order changes. & & \\
\path{thanks} & Responses to gratitude and positive feedback. & & \\
\path{greet} & Welcome messages for viewers entering the room. & & \\
\bottomrule
\end{tabularx}
\end{TRruntimeTable}

\fi

\subsection{Execution and Trajectory Record}
\label{app:execution_flow}

Figure~\ref{fig:system_overview} summarizes the implementation-level request path in a deliberately compact form. It separates the editable Harness modules from the repeated policy--tool loop and distinguishes resilience for an individual model call from recovery of the full trajectory. Product-specific tool names and Hook implementations are omitted because they vary across Harness versions.

% \begin{figure*}[tbp]
% \centering
% \includegraphics[width=\textwidth]{figures/harness_runtime_control_flow.pdf}
% \caption{Simplified Harness Agent runtime control flow. Prompt rules and dynamically loaded Skills assemble the model context, lifecycle Hooks validate inference, tool use, and stopping. A failed stop gate starts another round, while a passed gate produces the final response and structured trajectory. Per-call fallback and trajectory-level recovery protect different failure scopes.}
% \label{fig:harness_runtime_control_flow}
% \end{figure*}

A request traverses the following runtime stages:
\begin{enumerate}[leftmargin=*,itemsep=2pt]
\item \textbf{Load and normalize context.} Validate the request and assemble product, FAQ, room, and dialogue state.
\item \textbf{Assemble the prompt.} Combine the base instruction, active product and room configuration, relevant retrieved context, and any loaded Skills.
\item \textbf{Run the agent loop.} Under a configurable round limit, the policy can load Skills, issue one or more tool calls, incorporate their results, and propose a final reply.
\item \textbf{Apply Hook checks.} Lifecycle checks cover input validity, tool parameters, Skill-loading requirements, response format, known factuality risks, and retry control. The complete deployed Hook manifest remains version-bound rather than hard-coded in this description.
\item \textbf{Extract and dispatch the reply.} Remove internal reasoning fields, retain the structured trajectory, and route the public or private response to its downstream channel.
\end{enumerate}

\begin{definitionbox}{Example trajectory record}
{\scriptsize\ttfamily\raggedright
input.danmaku = ``How much is product \#3?''\\
context.current\_link\_id = 1\\
call[1] = get\_product\_info\_by\_link\_id(link\_id=3)\\
result[1] = \{name: ``Sunscreen SPF50'', price: 89\}\\
call[2] = load\_skill(skill=``item\_qa'')\\
reply = ``Link \#3 is Sunscreen SPF50, currently priced at 89 yuan.''\par
}
\end{definitionbox}

% Accuracy is binary, whereas Effectiveness admits scores of 0, 0.5, and 1; AVG is computed per sample before aggregation and rounding. The evaluator roles, majority-vote boundary, and clustered analysis are specified in Appendix~\ref{app:evaluation_details} rather than duplicated here.

\endgroup

\FloatBarrier
% 3. Method
\section{Harness-Aware Training (HAT)}
\label{sec:method}

\begin{insightbox}[Why Harness Evolution changes the training problem?]
Every promoted harness version creates a new state $h_{t+1}$ while the model remains fixed. A model that memorizes $h_t$ therefore becomes stale precisely when the system evolves successfully.
\end{insightbox}

\begin{figure*}[t]
\centering
\includegraphics[width=\textwidth]{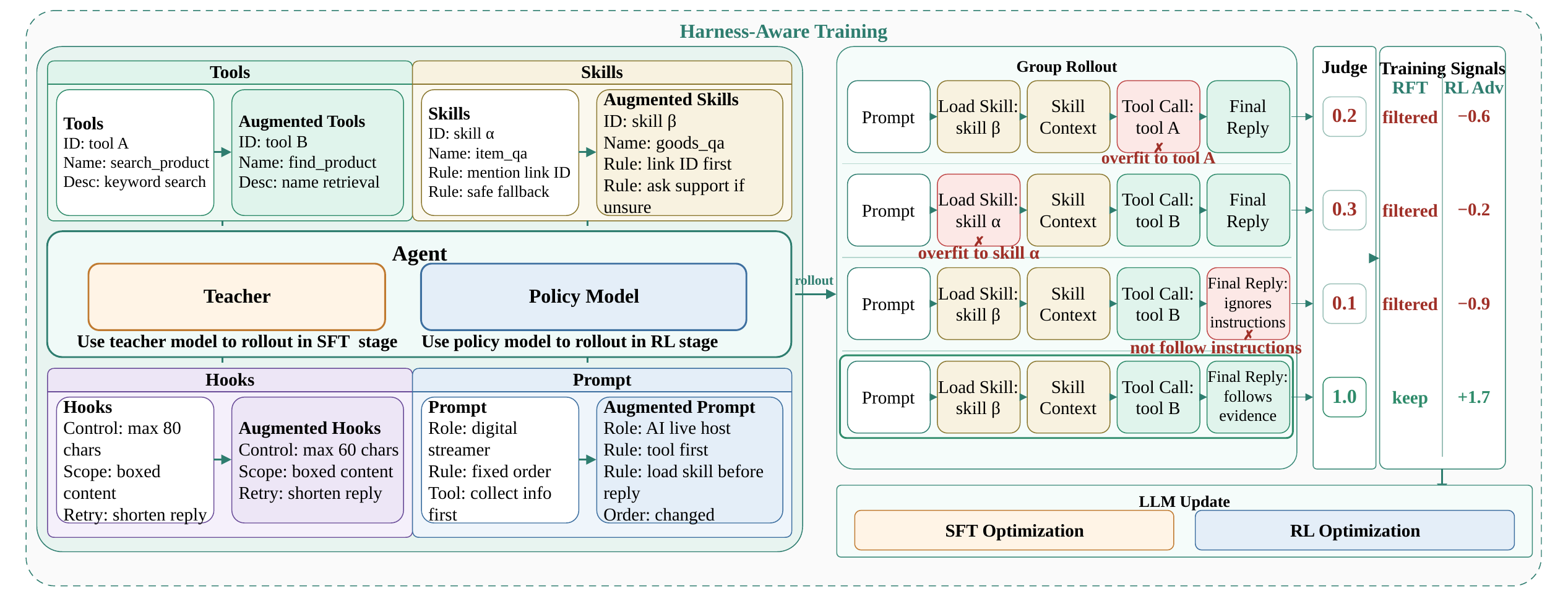}
\caption{Overview of harness-state augmentation and judging.
\emph{Left:} Examples of augmenting tools, skills, hooks, and prompts. The augmented harness is shared across training stages: a teacher model generates candidate trajectories for SFT, while the policy model produces rollouts for RL.
\emph{Right:} Representative RL trajectories. From top to bottom, Trajectory~1 invokes Tool~A, which is unavailable after augmentation, and is penalized by \emph{Tool Rationality}; Trajectory~2 selects a nonexistent Skill~$\alpha$ and is penalized by \emph{Skill Selection}; Trajectory~3 violates the modified prompt and is penalized by \emph{Accuracy}; and Trajectory~4 correctly adapts to the augmented harness and is rewarded. During SFT, the same judges guide rejection sampling. Together, augmentation and judging teach the model to understand the current harness rather than overfit to a fixed configuration.}
\label{fig:hsa_rl}
\end{figure*}

\subsection{Problem Formulation}
\label{sec:formulation}

We model a single agent interaction as $\pi_\theta(a \mid x, h)$, where $x$ denotes the user input together with its business context (e.g., live-streaming room state, current product listing), and $h \in \mathcal{H}$ is the \emph{Harness state}, the full configuration that governs agent behavior at inference time. Concretely,
\begin{equation}
h = (\mathcal{S},\; \mathcal{T},\; \mathcal{P},\; \mathcal{K}),
\label{eq:harness_state}
\end{equation}
where $\mathcal{S}$ is the set of active Skills (pluggable policy modules with reply rules and examples), $\mathcal{T}$ is the tool registry (schemas, parameter definitions, and descriptions), $\mathcal{P}$ is the system prompt assembled by the dynamic pipeline, and $\mathcal{K}$ is the set of Hooks (checkpoint logic for validation, retry, and format enforcement).

In production, $h$ evolves continuously: operations teams add Skills weekly, tool interfaces are updated as upstream services change, and prompts are revised in response to emerging bad cases. A model trained at time $t$ observes $h_t$ during training but faces $h_{t+1}, h_{t+2}, \ldots$ after deployment.

Our training objective is twofold: (1) improve policy effectiveness when deployment draws a new harness $h' \sim \mathcal{E}_{\textrm{deploy}}$ from a specified change envelope that is broader than any single training snapshot, (2) retain general-purpose capabilities such as instruction following after domain-specific fine-tuning.

\subsection{Harness-State Augmentation (HSA)}
\label{sec:augmentation}

The method is proposed to solve \emph{surface-form overfitting}: when training data contains a single, fixed harness, the objective permits shortcuts such as matching skill names by string similarity, selecting tools from memorized schema text, or following instructions only in a familiar template. The core hypothesis is that changing the surface form of the harness makes it harder for the model to rely on fixed names or templates as shortcuts. If a skill is renamed, its description rewritten, and distractors injected, functional information about the skill becomes more useful for routing than memorizing a single identifier.

We perturb the harness along five dimensions, each targeting a distinct aspect of surface-form dependence:

\begin{itemize}[leftmargin=*,nosep]
\item \emph{Skill Identifier:} We expand the skill set with synthetic plausible skills and noise skills, randomly mask subsets of existing skills, rename skills, and rewrite skill descriptions by a strong language model. These changes reduce the usefulness of a fixed skill identifier.

\item \emph{Skill Content:} Within each skill, we paraphrase individual rules, randomly mask a subset, and reorder them. This variation is intended to reduce reliance on rule position and exact wording.

\item \emph{Tool Definition:} We rename tools, rewrite descriptions. This transformation makes name matching less reliable while preserving the tool's function.

\item \emph{System Prompt:} We reorder top-level instruction blocks and items within blocks, and perturb numeric constraints (e.g., response length limits, maximum interaction rounds) within operationally valid ranges.

\item \emph{Hook:} We design controlled variants of retry behavior, message structure, and hook-triggered modifications to simulate the changes that the model may encounter after hooks are updated in production.
\end{itemize}

\subsection{Simulated On-Policy Environment}
\label{sec:simulation}

HSA addresses environment coverage but operates on \emph{offline} demonstration data: the model learns from teacher-generated trajectories that always follow the optimal path. In real deployment, however, the agent must handle situations that never appear in demonstrations. For example, a tool call returns an unexpected error, a Hook intercepts and forces a retry, or the agent selects a wrong Skill and must recover mid-conversation. Offline data provide no learning signal from the model's own failed trajectories and recovery attempts, because they contain only teacher trajectories. 

Moreover, the Harness Agent operates as a multi-step system: within a single user request, the model may perform Skill selection, tool invocation, result interpretation, and response generation across multiple reasoning rounds, with Hooks potentially triggering retries at each stage. Learning to navigate this dynamic interaction loop requires the model to actually \emph{experience} it during training, observing the consequences of its own actions rather than imitating a teacher's. Figure~\ref{fig:hsa_rl} illustrates the HSA-RL training loop, where augmented Harness states define the rollout environment, sampled agent trajectories expose skill, tool, and instruction-following errors, and trajectory feedback drives the policy update.

\paragraph{Environment Design}
We construct a production-informed live-streaming simulator that implements the control-flow elements needed for training: skills, tool executors, hooks, and bounded multi-round interaction. The simulator consists of four components:

\begin{itemize}[leftmargin=*,nosep]

\item \emph{Live-Streaming Input Simulation:}
The environment generates realistic user inputs by sampling from a pool of live-streaming scenarios: product inquiries, casual chat, purchasing intent, after-sales questions, and mixed-intent bullet comments. Each input is contextualized with a simulated room state including current product listings, historical conversation turns, and viewer action signals (e.g., adding to cart, joining the room).

\item \emph{Harness Agent Scheduling:}
The simulator implements the full Harness Agent control flow. Given the model's output at each reasoning step, the environment executes Skill routing logic (determining which Skills are loaded based on the model's selection), triggers Hooks at the appropriate checkpoints (input validation, parameter checking, response format verification, Skill-loading enforcement), and manages the multi-round interaction loop with a configurable maximum round limit.

\item \emph{Tool Execution Simulation:}
When the model invokes a tool, the simulator executes the call against sandboxed service replicas that return product details, pricing information, and inventory status from production catalog snapshots. The simulator also injects controlled failures such as timeout errors, malformed responses, and permission denials at calibrated rates to expose the model to error-recovery scenarios.

\item \emph{Augmented Harness Configuration:}
The simulated environment operates under augmented Harness states produced by the method described in \S\ref{sec:augmentation}. This aligns the offline and on-policy stages to the same declared change envelope.

\end{itemize}
%it does not establish robustness to perturbation families absent from that envelope.

\subsection{Training Pipeline}
\label{sec:pipeline}

The complete training pipeline consists of three stages, each addressing a distinct capability gap:

\begin{figure*}[t]
\centering
\includegraphics[width=\textwidth]{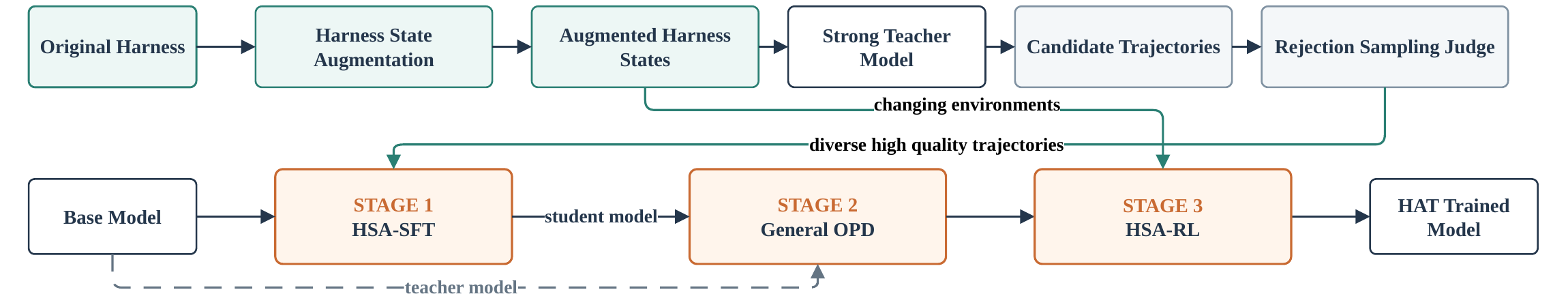}
\caption{Overview of Harness-Aware Training (HAT). Harness-State Augmentation (HSA) transforms the original Harness into diverse, task-preserving states. A strong teacher generates candidate trajectories under these states, which are filtered by a rejection-sampling judge to provide diverse, high-quality supervision for HSA-SFT. Starting from the base model, HAT then follows three stages: HSA-SFT on the filtered trajectories; General OPD, using the base model as the teacher and the HSA-SFT checkpoint as the student to mitigate general-capability degradation; and HSA-RL in changing HSA environments. The resulting policy is the final HAT-trained model.}
\label{fig:harness_aware_training_design}
\end{figure*}

\paragraph{Stage 1: SFT with Harness-State Augmentation (HSA-SFT)}
HSA-SFT constructs supervised fine-tuning samples from real live-room interactions. Each original sample contains the viewer query, live-room context, and product context, and is paired with a corresponding Harness which includes the available skills, tool definitions, system prompt, and Hook configuration. To reduce the model's dependence on the surface form of a single harness, we first apply HSA to the Harness base state in Figure~\ref{fig:harness_aware_training_design} before generating supervised trajectories. This produces both the original Harness and multiple augmented Harness variants. HSA changes skill names and descriptions, skill-rule order, tool descriptions, system-prompt structure, and the surface form of Hook messages and retry behavior.

The teacher model then generates candidate trajectories under both the original harness and the augmented harness variants, so the supervised signals cover multiple simulated environments~\citep{wu2026beyond}. These candidate trajectories are filtered according to Accuracy and Effectiveness. As a result, supervised learning exposes the student to the same task semantics under different harness realizations, encouraging it to use the functional meaning of the current Harness rather than fixed names or template matches.

Hooks in production may inform the model about formatting errors, parameter errors, or invalid outputs, and ask it to revise. We want the model to correct itself when receiving a hook reminder, but we do not want it to learn a default strategy of producing an erroneous output first and relying on retries. Therefore, for most trajectories that contain hook-triggered retries, we remove the intermediate failed outputs and hook interactions and keep only the final correct behavior. We retain a small fraction of complete retry trajectories so that the model still learns how to interpret hook reminders and recover after a failure.

\paragraph{Stage 2: General Domain On-Policy Distillation (General OPD)}

Motivated by prior practice showing that On-Policy Distillation can improve general instruction-following ability~\citep{lu2025onpolicydistillation}, we add a General OPD stage after HSA-SFT. We use the pre-HSA-SFT base model as the teacher and perform on-policy distillation on the Tulu3~\citep{lambert2025tulu3pushingfrontiers} general instruction dataset. The student samples responses on-policy for Tulu inputs and is trained by minimizing the KL divergence between the student distribution and the base-model distribution:
\begin{equation}
\mathcal{L}_{\textrm{OPD}} = E_{x \sim \mathcal{D}_{\textrm{gen}}} \big[\, D_{\textrm{KL}}\!\left( p_\theta(\cdot \mid x) \,\|\, p_{\textrm{base}}(\cdot \mid x) \right) \,\big].
\label{eq:opd}
\end{equation}

\paragraph{Stage 3: Agentic RL with Harness-State Augmentation (HSA-RL)}

From the checkpoint trained by HSA-SFT and General OPD, we further optimize the policy in the simulated on-policy environment. This environment uses the production-informed simulator defined in Section~\ref{sec:simulation}, including live-streaming scenarios, multi-round interaction, skill loading, tool execution, and Hooks. In HSA-RL, HSA is applied before sequence generation: before each rollout, we first decide whether the current sample uses the original Harness or an augmented Harness, and the model then generates a complete interaction trajectory under that Harness. In this way, part of the rollouts are exposed to the original production configuration, while the rest are exposed to domain-preserving augmented Harness states, maintaining adaptation to the original environment while expanding coverage of Harness changes. For optimization, we use the Agentic RL objective described below, with the complete training procedure summarized in Algorithm~\ref{alg:agentic_rl}.

Within the simulated environment, we optimize the policy based on Group Relative Policy Optimization (GRPO)~\citep{shao2024deepseekmath}. For each prompt, $G$ complete trajectory rollouts (group size) are sampled and split into tool segments and reply segments. The reward signal is computed from 4 dimensions:
\begin{itemize}[leftmargin=*,nosep]
\item \emph{Accuracy}: whether the final response contains factually correct information and follows the instructions defined in prompts or skills. Applied on reply state;
\item \emph{Effectiveness}: whether the response adequately addresses the user's intent with appropriate detail and actionability. Applied on reply state;
\item \emph{Tool Rationality}: whether the tool invocation chain is logically sound and efficiently structured (no redundant calls, correct parameter usage), this reward can prevent model from overfitting to specific tools. Applied on tool state;
\item \emph{Skill Selection}: whether the model correctly identifies the user's intent and the current context to load the appropriate Skills, this reward can prevent model from overfitting to specific skills. Applied on tool state.
\end{itemize}

As an auxiliary regularizer, we discourage unnecessarily long chain-of-thought (CoT) spans~\citep{zhang2026tokensqueeze,pan2026chainpruneevaluatingreducingredundancy}. For state group $g$, let $L_i^g$ be the sum of tokens in the CoT spans extracted from trajectory $i$. Tool-call and final-reply states are measured separately, so one state does not consume the other's length allowance. We define
\begin{equation}
\begin{aligned}
r_\text{CoT}(L)&=
\begin{cases}
0, & L \leq L_{\mathrm{low}},\\
\dfrac{L-L_{\mathrm{low}}}{L_{\mathrm{high}}-L_{\mathrm{low}}},
& L_{\mathrm{low}} < L < L_{\mathrm{high}},\\
1, & L \geq L_{\mathrm{high}},
\end{cases}
\end{aligned}
\label{eq:cot_penalty}
\end{equation}
Here $L_{\mathrm{low}}$ and $L_{\mathrm{high}}$ are tunable no-penalty and saturation thresholds, respectively, with $L_{\mathrm{low}}<L_{\mathrm{high}}$. We set them to 100 and 200 tokens in the reported experiments. Thus the raw contribution ranges from 0 to $-0.1$ per state. This component is intended to reduce avoidable reasoning length.

The tool and reply segments are treated as separate groups ($g \in \mathcal{G} = \{\texttt{tool}, \texttt{reply}\}$) for advantage computation. Since each group carries multiple reward dimensions, we adopt \emph{Group reward Decomposed Policy Optimization} (GDPO)~\citep{liu2026gdpogrouprewarddecouplednormalization} to estimate advantages: for each task reward dimension $d$ within state group $g$, a group-normalized advantage $\hat{A}^{(d)}_i = (r^{(d)}_i - \mu^{(d)}_g)/\sigma^{(d)}_g$ is computed. The CoT score is normalized as its own dimension, $\hat{A}^{(\mathrm{CoT},g)}_i=(r_\text{CoT}(L_i^g)-\mu_g^{(\mathrm{CoT})})/\sigma_g^{(\mathrm{CoT})}$, and its negative weight is applied after this per-dimension normalization:
\begin{equation}
\hat{A}_i^g = \sum_{d \in \mathcal{D}_g}\hat{A}_i^{(d)}
-\lambda\hat{A}^{(\mathrm{CoT},g)}_i,
\qquad
\hat{A}_i^{g,\star}=\frac{\hat{A}_i^g-\mu^g}{\sigma^g}.
\label{eq:gdpo_cot}
\end{equation}
Here $\mu$ and $\sigma$ denote the corresponding within-group means and standard deviations, and all four task dimensions retain unit weight. For importance sampling, we adopt the sequence-level scheme of GSPO~\citep{zheng2025gspo}, giving the clipped policy gradient objective:
\begin{equation}
\begin{aligned}
\mathcal{L} &= -\frac{1}{|\mathcal{G}|G}\sum_{g}\sum_{i}
\min\!\Bigl(
  \rho_i(\theta)\,\hat{A}^{g,\star}_i,\;
  \mathrm{clip}(\rho_i(\theta), 1{\pm}\varepsilon)\,\hat{A}^{g,\star}_i
\Bigr), \\
\rho_i(\theta) &= \left(\frac{\pi_\theta(y_i \mid x)}{\pi_{\theta_{\mathrm{old}}}(y_i \mid x)}\right)^{\!\frac{1}{|y_i|}}, \\
&= \exp\!\left(\frac{1}{|y_i|}\sum_{t=1}^{|y_i|}\log\frac{\pi_\theta(y_{i,t}\mid x, y_{i,<t})}{\pi_{\theta_{\mathrm{old}}}(y_{i,t}\mid x, y_{i,<t})}\right).
\end{aligned}
\label{eq:rl_obj}
\end{equation}
Here $\pi_\theta$ and $\pi_{\theta_{\mathrm{old}}}$ are the current and reference policies; $y_i$ is the generated sequence of length $|y_i|$; $y_{i,t}$ is the $t$-th token and $y_{i,<t}$ its prefix; $x$ is the input context; and $\varepsilon$ is the PPO clip range~\citep{schulman2017ppo}. The sequence-level ratio $\rho_i(\theta)$ replaces token-level importance sampling to stabilize training under the sparse expert routing of the MoE architecture.

When an individual trajectory has zero advantage after GDPO normalization, it provides no gradient signal while diluting the effective batch. Such zero-gradient samples are discarded before optimization~\citep{yu2025dapoopensourcellmreinforcement}. Algorithm~\ref{alg:agentic_rl} gives the complete HSA-RL procedure.

\ifdefined\TRInlineSupplementarySections
\begin{algorithm}[!htbp]
\caption{Agentic RL with Harness-State Augmentation (HSA-RL)}
\label{alg:agentic_rl}
\scriptsize
\begin{algorithmic}[1]
\Require Policy $\pi_\theta$; group size $G$; state groups $\mathcal{S}=\{\texttt{tool},\texttt{reply}\}$; reward dims $\mathcal{D}_{\texttt{tool}}$ and $\mathcal{D}_{\texttt{reply}}$; CoT weight $\lambda$; clip range $\varepsilon$
\For{each training iteration}
  \State Sample a mini-batch of queries and augmented Harness states $(x,h)$
  \State $\pi_{\theta_{\mathrm{old}}} \gets \pi_\theta$; initialize $\mathcal{L}_{\texttt{tool}}\gets0$, $\mathcal{L}_{\texttt{reply}}\gets0$
  \For{each $(x,h)$ in the mini-batch}
    \For{$i=1,\ldots,G$}
      \State Roll out trajectory $\tau_i$ under $\pi_{\theta_{\mathrm{old}}}(\cdot\mid x,h)$
      \State Split $\tau_i$ into tool segment $y_i^{\texttt{tool}}$ and reply segment $y_i^{\texttt{reply}}$
      \State Create masks $m_i^{\texttt{tool}}$ and $m_i^{\texttt{reply}}$ over generated action tokens; mask prompts and observations
      \State Score $\{r_i^{(d)}\}_{d\in\mathcal{D}_{\texttt{tool}}}$ from tool behavior and $\{r_i^{(d)}\}_{d\in\mathcal{D}_{\texttt{reply}}}$ from final-reply quality
      \State Extract CoT lengths $L_i^{\texttt{tool}}$ and $L_i^{\texttt{reply}}$ and compute $q_i^s\gets q(L_i^s)$ for $s\in\mathcal{S}$
    \EndFor
    \For{each state group $s\in\{\texttt{tool},\texttt{reply}\}$}
      \For{each reward dimension $d\in\mathcal{D}_s$}
        \State $\hat{A}_i^{(d)} \gets (r_i^{(d)}-\mu_s^{(d)})/\sigma_s^{(d)}$ for $i=1,\ldots,G$
      \EndFor
      \State $\hat{A}_i^{(\mathrm{CoT},s)}\gets(q_i^s-\mu_s^{(\mathrm{CoT})})/\sigma_s^{(\mathrm{CoT})}$
      \State $\hat{A}_i^s\gets\sum_{d\in\mathcal{D}_s}\hat{A}_i^{(d)}-\lambda\hat{A}_i^{(\mathrm{CoT},s)}$; normalize to $\hat{A}_i^{s,\star}$
      \For{each sequence $i=1,\ldots,G$ with $\hat{A}_i^{s,\star}\ne0$}
        \State $\rho_{i,t}^s(\theta)\gets\dfrac{\pi_\theta(y_{i,t}^s\mid y_{i,<t}^s,x,h)}{\pi_{\theta_{\mathrm{old}}}(y_{i,t}^s\mid y_{i,<t}^s,x,h)}$
        \State $\ell_{i,t}^s\gets-\min\bigl(\rho_{i,t}^s\hat{A}_i^{s,\star},\mathrm{clip}(\rho_{i,t}^s,1{-}\varepsilon,1{+}\varepsilon)\hat{A}_i^{s,\star}\bigr)$
        \State $\mathcal{L}_s\gets\mathcal{L}_s+\dfrac{\sum_t m_{i,t}^s\ell_{i,t}^s}{\sum_t m_{i,t}^s}$
      \EndFor
    \EndFor
  \EndFor
  \State Update $\theta$ by minimizing the mean of $\mathcal{L}_{\texttt{tool}}$ and $\mathcal{L}_{\texttt{reply}}$
\EndFor
\end{algorithmic}
\end{algorithm}

\fi

\FloatBarrier
% 4. Evaluation
\section{Evaluation}
\label{sec:evaluation}
This section describes the evaluation, the scoring metrics for each source, and the deployment replay protocol used to measure actual latency and serving performance.
\subsection{Evaluation Sources}
\label{sec:eval_sets_sources}

We organize the evaluation around the requirements in the live streaming application. For offline evaluation, Live-Stream QA and Harness-Variant QA are both built from real live-room interactions and cover the main industry scenarios in our application domain, with the available internal intent mix summarized in Appendix Table~\ref{tab:scenario_dist}. Live-Stream QA tests industrial reply quality, while Harness-Variant QA tests generalization and robustness to changing Harness contexts. We also use a separate real live-room calibration cohort to align the Harness-based Agent-as-a-Judge with human labels. To test broader tool and prompt robustness, we construct synthetic Tool Robustness and Prompt Robustness subsets by expanding real-industrial scenario seeds with LLM generation and consistency validation, as shown in Appendix~\ref{app:t3_robustness}. We also evaluate general instruction-following ability on the public IFEval benchmark~\citep{zhou2023instruction}. To test actual deployment latency and serving performance, we use a fixed complete-Agent deployment replay. Table~\ref{tab:eval_contract} summarizes the size, source, and role of each evaluation source.

\begin{table}[t]
\centering
\caption{Evaluation sets and sources. The sets cover real live-stream reply quality, in-family Harness variation, synthetic tool and prompt robustness, general instruction following, and actual deployment latency.}
\label{tab:eval_contract}
\footnotesize
\renewcommand{\arraystretch}{1.08}
\setlength{\tabcolsep}{3.5pt}
\begin{tabularx}{\linewidth}{@{}>{\raggedright\arraybackslash}p{3.80cm}>{\raggedright\arraybackslash}p{0.80cm}l@{\hspace{5em}}>{\raggedright\arraybackslash}X@{}}
\toprule
\tablehead{Set} & \tablehead{Size} & \tablehead{Source} & \tablehead{Evidentiary role} \\
\midrule
$\mathcal{T}_1$ Live-Stream QA & 978 & Real live-room w/ fixed Harness & Primary industrial quality (live-stream reply). \\
$\mathcal{T}_2$ Harness-Variant QA & 978 & Real live-room w/ augmented Harness & In-family robustness to new Harness contexts. \\
$\mathcal{T}_3$ Synthetic Live-Stream QA & 2023 & Synthetic live-room scenarios & In-domain transfer under broader tools/prompts. \\
$\mathcal{T}_4$ IFEval & 541 & Public Benchmark & General instruction following (official evaluator). \\
$\mathcal{C}_{\mathrm{judge}}$ & 482 & Real live-room, human-labeled & Judge calibration, not a policy test set. Also dev-set for harness evolving. \\
$\mathcal{D}_{\mathrm{perf}}$ & 110 & Deployment replay  & End-to-end deployment latency. \\
\bottomrule
\end{tabularx}
\end{table}

\subsection{Metrics and Scoring}
\label{sec:metrics_scoring}

For Live-Stream QA and Harness-Variant QA, we evaluate each response with Accuracy and Effectiveness using the Harness-based Agent-as-a-Judge, following the LLM-as-a-Judge paradigm~\citep{zheng2024judging}. Appendix~\ref{app:judge_roles}. Accuracy is a binary metric that penalizes factual errors, unsupported product claims, false promises, and capability overreach. Effectiveness measures whether the response addresses the viewer's intent and provides useful information, with scores of 0, 0.5, or 1. We report AVG as the per-sample arithmetic mean of Accuracy and Effectiveness, computed before rounding.

For Tool and Prompt Robustness, two subsets of Synthetic Live-Stream QA, we use task-specific rubric metrics. The rubrics are generated with the synthetic task simutanuously. Tool Robustness uses ACQ, RC, and TC: ACQ measures whether the response \textbf{A}ddresses the \textbf{C}ore viewer \textbf{Q}uery, RC measures whether the \textbf{R}esponse \textbf{C}ontains the required information returned by tools, and TC measures whether the expected \textbf{T}ool is \textbf{C}alled. Prompt Robustness uses ACQ, FI, and TCO: ACQ keeps the same core-query definition, FI evaluates whether the response \textbf{F}ollows the active \textbf{I}nstructions, and TCO checks whether \textbf{T}ool \textbf{C}alls follow the required \textbf{O}rder.

For IFEval, we use its official evaluator and report prompt-level and instruction-level accuracy (abbreviated as IFE-P and IFE-I hereafter). IFE-P measures whether the model satisfies the full prompt-level instruction set, while IFE-I measures whether individual instructions are followed.

For deployment replay, we report wall-clock latency P50 and P95, TTFT, decoding throughput, execution success, and 15-second attainment. Wall-clock latency is the time from receiving the input query to obtaining the final Agent answer. TTFT is the time to the first token. Execution success indicates completion of the Agent request, not answer quality. The 15-second attainment denominator includes all measured requests.

\subsection{Deployment Inference Test}
\label{sec:deployment_benchmark}

For deployment inference performance testing, we use 110 cases $\mathcal{D}_{\mathrm{perf}}$, mark 10 as warm-up, and compute metrics over the remaining 100 requests. Every request executes the complete Agent with the same Harness commit, real MCP services, at most four agent loops, and a 5-second tool timeout. Generation uses temperature 1.0 and top-$p$ 0.95. Qwen3.6-35B-A3B and our HAT trained model run in BF16 on one H20 with TP=PP=1. Their NextN MTP modules supply draft tokens, which the serving engine processes through the same EAGLE-style speculative generation and verification path with three speculative steps and four draft tokens. Appendix~\ref{app:deployment_inference} gives the complete engine configuration.

\FloatBarrier
% 5. Experiments - original content + Blind Test + Deployment
\def\TRCompactExperimentPagination{1}
% \onecolumn
\thispagestyle{fancy}
\section{Experiments}
\label{sec:experiments}

\subsection{Experimental Setup}
\label{sec:exp_setup}

The training experiments are conducted on Qwen3.6-35B-A3B. The training data are collected from real live-streaming scenarios, with 10K examples for SFT and 4K interaction tasks for RL. During evaluation, Qwen3.6-35B-A3B and our trained checkpoints are deployed on one NVIDIA H20, while the Top models are accessed through Bailian API. We evaluate the checkpoints on the offline sets and deployment replay defined in Section~\ref{sec:evaluation}. For $\mathcal{T}_1$--$\mathcal{T}_3$, all model comparisons use the same Harness; for $\mathcal{T}_4$, all models are evaluated with the same official IFEval code.

%% ============================================================
% Keep the dedicated results page in the two-column paper, but let the
% single-column technical report use the remaining space on the current page.
\ifdefined\TRCompactExperimentPagination\else
\clearpage
\fi
\subsection{Main Results}
\label{sec:main_results}

Table~\ref{tab:main_results} compares the models in terms of business response quality, robustness to Harness changes, and general instruction following. HAT achieves AVG scores of 94.8 on $\mathcal{T}_1$ and 94.6 on $\mathcal{T}_2$, exceeding the best Top-model scores of 93.0 and 93.5 under the same evaluation protocol. The difference between its performance on the original and Harness-variant settings is only 0.2 points, compared with 4.9 points for Base and 1.3 points for Fixed-Harness SFT. This result indicates that HAT maintains stable business performance across the evaluated Harness changes.
Fixed-Harness SFT improves domain performance but introduces clear generalization degradation. On $\mathcal{T}_2$, it substantially improves Effectiveness, while Accuracy decreases from 88.7 to 86.9. Its Prompt Robustness AVG also drops from 72.8 to 68.2, as instruction following and tool-call ordering deteriorate despite better coverage of the core query. In contrast, HAT reaches AVG scores of 84.0 and 77.6 on Tool and Prompt Robustness. It also achieves 83.5 IFE-P and 88.7 IFE-I, avoiding the 7.7 and 5.3 point drops caused by Fixed-Harness SFT. Overall, HAT provides a better balance among domain performance, robustness to Harness changes, and general capability preservation.

\begin{strip}
\vspace{2pt}
\captionsetup{type=table,hypcap=false,font={footnotesize,sf}}
\caption{Main results across offline evaluation sets. $\mathcal{T}_1$ measures real live-stream reply quality, $\mathcal{T}_2$ measures robustness to Harness-Variant QA, $\mathcal{T}_3$ measures Tool and Prompt Robustness on synthetic live-streaming digital avatar scenarios, and $\mathcal{T}_4$ measures general instruction following with IFEval. Acc, Eff, and AVG denote Accuracy, Effectiveness, and their per-sample average; ACQ, RC, TC, FI, and TCO are the rubric metrics defined in Section~\ref{sec:metrics_scoring}. Fixed-Harness SFT denotes naive supervised fine-tuning without HSA, while HAT (Ours) denotes the model obtained by the full HSA-SFT + General OPD + HSA-RL pipeline.}
\label{tab:main_results}

\footnotesize
\renewcommand{\arraystretch}{0.92}
\setlength{\tabcolsep}{4.5pt}
\setlength{\aboverulesep}{2pt}
\setlength{\belowrulesep}{2pt}
\noindent{\sffamily\bfseries\color{TableAccent}Panel A \textcolor{TableMuted}{· Live-Stream QA quality and general instruction following}}\par\vspace{2pt}
\begin{tabularx}{\textwidth}{@{}l *{8}{>{\centering\arraybackslash}X}@{}}
\toprule
\multirow{2}{*}{\tablehead{Method}} & \multicolumn{3}{c}{$\mathcal{T}_1$ Live-Stream QA} & \multicolumn{3}{c}{$\mathcal{T}_2$ Harness-Variant QA} & \multicolumn{2}{c}{$\mathcal{T}_4$ IFEval} \\
\cmidrule(lr){2-4} \cmidrule(lr){5-7} \cmidrule(lr){8-9}
 & Acc & Eff & AVG & Acc & Eff & AVG & IFE-P & IFE-I \\
\midrule
\rowcolor{TableGroup}\multicolumn{9}{@{}l}{\sffamily\bfseries\color{TableAccent}Top models \normalfont\itshape\color{TableMuted}} \\
\quad DeepSeek-V4-pro & 93.7 & 88.7 & 91.2 & 92.1 & 89.1 & 90.6 & 88.2 & 92.0 \\
\quad GLM-5.2 & 93.0 & 93.0 & 93.0 & 93.6 & 93.4 & 93.5 & 89.6 & 92.7 \\
\quad DeepSeek-V4-flash & 90.7 & 92.0 & 91.4 & 91.3 & 90.6 & 91.0 & 90.0 & 93.1 \\
\quad Qwen3.7-max & 93.8 & 91.2 & 92.5 & 93.2 & 92.5 & 92.8 & 90.4 & 93.6 \\
\quad Qwen3.6-plus & 91.7 & 85.6 & 88.7 & 93.3 & 89.3 & 91.3 & 87.2 & 91.2 \\
\midrule
\rowcolor{TableGroup}\multicolumn{9}{@{}l}{\sffamily\bfseries\color{TableAccent}Qwen3.6-35B-A3B \normalfont\itshape\color{TableMuted}} \\
\quad Base & 86.5 & 74.1 & 80.3 & 88.7 & 62.1 & 75.4 & 81.5 & 87.7 \\
\quad Fixed-Harness SFT & 90.0 & 88.9 & 89.5 & 86.9 & 89.5 & 88.2 & 73.8\textsuperscript{\scriptsize$\downarrow$7.7} & 82.4\textsuperscript{\scriptsize$\downarrow$5.3} \\
\rowcolor{TableHighlight}\quad \textbf{HAT (Ours)} & 95.8 & 93.8 & 94.8 & 94.5 & 94.7 & 94.6 & 83.5\textsuperscript{\scriptsize$\uparrow$2.0} & 88.7\textsuperscript{\scriptsize$\uparrow$1.0} \\
\bottomrule
\end{tabularx}

\vspace{4pt}
\noindent{\sffamily\bfseries\color{TableAccent}Panel B \textcolor{TableMuted}{· Broad AI-generated live-stream avatar scenarios}}\par\vspace{2pt}
\begin{tabularx}{\textwidth}{@{}l *{8}{>{\centering\arraybackslash}X}@{}}
\toprule
\multirow{2}{*}{\tablehead{Method}} & \multicolumn{4}{c}{$\mathcal{T}_3$ Tool Robustness} & \multicolumn{4}{c}{$\mathcal{T}_3$ Prompt Robustness} \\
\cmidrule(lr){2-5} \cmidrule(lr){6-9}
 & ACQ & RC & TC & AVG & ACQ & FI & TCO & AVG \\
\midrule
\rowcolor{TableGroup}\multicolumn{9}{@{}l}{\sffamily\bfseries\color{TableAccent}Top models \normalfont\itshape\color{TableMuted}} \\
\quad DeepSeek-V4-pro & 98.7 & 74.2 & 92.4 & 84.5 & 98.2 & 80.8 & 80.9 & 83.4 \\
\quad GLM-5.2 & 96.9 & 74.6 & 96.7 & 86.7 & 98.8 & 84.8 & 74.7 & 84.0 \\
\quad DeepSeek-V4-flash & 98.2 & 72.3 & 95.1 & 85.5 & 98.2 & 78.5 & 74.4 & 83.4 \\
\quad Qwen3.7-max & 98.0 & 63.0 & 96.2 & 81.6 & 95.3 & 85.2 & 65.0 & 81.1 \\
\quad Qwen3.6-plus & 98.9 & 61.5 & 94.4 & 80.0 & 95.9 & 79.0 & 54.4 & 74.7 \\
\midrule
\rowcolor{TableGroup}\multicolumn{9}{@{}l}{\sffamily\bfseries\color{TableAccent}Qwen3.6-35B-A3B \normalfont\itshape\color{TableMuted}} \\
\quad Base & 91.9 & 51.5 & 83.3 & 69.5 & 89.4 & 73.3 & 62.8 & 72.8 \\
\quad Fixed-Harness SFT & 92.3 & 66.6 & 94.5 & 82.0 & 92.9 & 65.5 & 60.6 & 68.2 \\
\rowcolor{TableHighlight} \quad \textbf{HAT (Ours)} & 99.1 & 70.1 & 94.7 & 84.0 & 99.4 & 78.0 & 65.3 & 77.6 \\
\bottomrule
\end{tabularx}

% \begin{insightbox}[Key Insights]
% \begin{itemize}[leftmargin=1.25em,nosep,itemsep=1.5pt]
% \item \textbf{Fixed-Harness SFT overfits.} Fixed-Harness SFT improves the in-family task scores but lowers $\mathcal{T}_3$ Prompt Robustness ($-$4.6) and $\mathcal{T}_4$ IFEval ($-$7.7/$-$5.3), suggesting that training on one harness version can encourage shortcut learning and reduce robustness and generalization.
% \item \textbf{HAT generalizes.} HAT trained model reaches 94.8 on $\mathcal{T}_1$, exceeding the strongest Top-model score of 93.0 on the same business dataset, while avoiding the robustness and generalization drops observed after Fixed-Harness SFT.
% \end{itemize}
% \end{insightbox}
\end{strip}

% Keep later floats from moving ahead of either panel of the main-results
% table.  This is especially important in the single-column technical report,
% where `strip` is intentionally a no-op.
\FloatBarrier

%% ============================================================
% \subsection{Ablation Studies}
% \label{sec:ablation}

\subsection{Effectiveness of Training Stages}

Table~\ref{tab:pipeline_ablation} reports cumulative training through each pathway's shared initialization, followed by alternative Agentic RL branches under the original Harness and HSA environments.

% In the single-column technical report, keep this table exactly beside its
% discussion so it cannot float between Panels A and B above.  The two-column
% paper still needs a full-width float and may place it at the next page top.
\ifdefined\TRCompactExperimentPagination
\begin{table}[H]
\else
\begin{table*}[!t]
\fi
\captionsetup{type=table,hypcap=false}
\caption{Training-pipeline ablation. Each row adds one stage to the previous checkpoint; $\uparrow$/$\downarrow$ denotes a descriptive score change from the row above and does not include retraining variance.}
\label{tab:pipeline_ablation}
\small
\setlength{\tabcolsep}{4.2pt}
\begin{tabular*}{\textwidth}{@{\extracolsep{\fill}}l cccccc@{}}
\toprule
\tablehead{Harness Configuration} & \tablehead{$\mathcal{T}_1$} & \tablehead{$\mathcal{T}_2$} & \tablehead{$\mathcal{T}_3$ Tool Robustness} & \tablehead{$\mathcal{T}_3$ Prompt Robustness} & \tablehead{$\mathcal{T}_4$ IFE-P} & \tablehead{$\mathcal{T}_4$ IFE-I} \\
\midrule
\rowcolor{TableGroup}\multicolumn{7}{@{}l}{\sffamily\bfseries\color{TableAccent}Non-augmented \normalfont\itshape\color{TableMuted}} \\
\quad Base & 80.3 & 75.4 & 69.5 & 72.8 & 81.5 & 87.7 \\
\quad +SFT & 89.5\textsuperscript{\scriptsize$\uparrow$9.2} & 88.2\textsuperscript{\scriptsize$\uparrow$12.8} & 82.0\textsuperscript{\scriptsize$\uparrow$12.5} & 68.2\textsuperscript{\scriptsize$\downarrow$4.6} & 73.8\textsuperscript{\scriptsize$\downarrow$7.7} & 82.4\textsuperscript{\scriptsize$\downarrow$5.3} \\
\quad +General OPD & 89.5 & 89.1\textsuperscript{\scriptsize$\uparrow$0.9} & 84.3\textsuperscript{\scriptsize$\uparrow$2.3} & 72.6\textsuperscript{\scriptsize$\uparrow$4.4} & 82.3\textsuperscript{\scriptsize$\uparrow$8.5} & 87.9\textsuperscript{\scriptsize$\uparrow$5.5} \\
\quad +RL & 95.1\textsuperscript{\scriptsize$\uparrow$5.6} & 94.4\textsuperscript{\scriptsize$\uparrow$5.3} & 83.7\textsuperscript{\scriptsize$\downarrow$0.6} & 66.7\textsuperscript{\scriptsize$\downarrow$5.9} & 82.7\textsuperscript{\scriptsize$\uparrow$0.4} & 87.8\textsuperscript{\scriptsize$\downarrow$0.1} \\
\midrule
\rowcolor{TableGroup}\multicolumn{7}{@{}l}{\sffamily\bfseries\color{TableAccent}Augmented \normalfont\itshape\color{TableMuted}} \\
\quad +HSA-SFT & 90.0\textsuperscript{\scriptsize$\uparrow$9.7} & 90.2\textsuperscript{\scriptsize$\uparrow$14.8} & 85.2\textsuperscript{\scriptsize$\uparrow$15.7} & 77.3\textsuperscript{\scriptsize$\uparrow$4.5} & 81.7\textsuperscript{\scriptsize$\uparrow$0.2} & 87.4\textsuperscript{\scriptsize$\downarrow$0.3} \\
\quad +General OPD & 90.9\textsuperscript{\scriptsize$\uparrow$0.9} & 90.9\textsuperscript{\scriptsize$\uparrow$0.7} & 85.6\textsuperscript{\scriptsize$\uparrow$0.4} & 77.1\textsuperscript{\scriptsize$\downarrow$0.2} & 81.9\textsuperscript{\scriptsize$\uparrow$0.2} & 88.0\textsuperscript{\scriptsize$\uparrow$0.6} \\
\rowcolor{TableHighlight}\quad +HSA-RL (\textbf{HAT}) & 94.8\textsuperscript{\scriptsize$\uparrow$3.9} & 94.6\textsuperscript{\scriptsize$\uparrow$3.7} & 84.0\textsuperscript{\scriptsize$\downarrow$1.6} & 77.6\textsuperscript{\scriptsize$\uparrow$0.5} & 83.5\textsuperscript{\scriptsize$\uparrow$1.6} & 88.7\textsuperscript{\scriptsize$\uparrow$0.7} \\
\bottomrule
\end{tabular*}
\ifdefined\TRCompactExperimentPagination
\end{table}
\else
\end{table*}
\fi

First, Fixed-Harness SFT substantially improves task-specific performance: it raises $\mathcal{T}_1$ and $\mathcal{T}_2$ by 9.2 and 12.8 points, respectively, and improves $\mathcal{T}_3$ Tool Robustness by 12.5 points. However, these gains come with clear generalization degradation: Prompt Robustness decreases by 4.6 points, while IFE-P and IFE-I drop by 7.7 and 5.3 points, respectively. This suggests that the model can overfit to the surface form of a fixed Harness. In contrast, HSA-SFT improves $\mathcal{T}_1$, $\mathcal{T}_2$, and Tool Robustness by 9.7, 14.8, and 15.7 points, respectively, while also increasing Prompt Robustness by 4.5 points and largely preserving the base model's IFEval performance. These results show that SFT across diverse, semantically equivalent Harness states can improve task-specific performance without sacrificing generalization.

Second, both General OPD and HSA-SFT produce models with strong generalization, and their benefits can be combined. Applying General OPD after Fixed-Harness SFT recovers 4.4 points on Prompt Robustness and 8.5/5.5 points on IFE-P/IFE-I, while further improving $\mathcal{T}_2$ and Tool Robustness. HSA-SFT, in contrast, prevents the generalization degradation caused by Fixed-Harness SFT from the outset. Adding General OPD after HSA-SFT further raises both $\mathcal{T}_1$ and $\mathcal{T}_2$ to 90.9 and improves Tool Robustness and both IFEval metrics, while keeping Prompt Robustness nearly unchanged ($-0.2$). General OPD and HSA-SFT are therefore complementary: the former recovers general capabilities, whereas the latter improves generalization across Harness states, and their combination yields better overall performance than either alone.

Finally, the RL stage consistently improves performance on the business evaluation sets, regardless of whether HSA is used. Along the non-augmented pathway, RL improves $\mathcal{T}_1$ and $\mathcal{T}_2$ by 5.6 and 5.3 points, respectively; along the HSA pathway, HSA-RL provides corresponding gains of 3.9 and 3.7 points. The two pathways, however, differ substantially in Prompt Robustness. Without HSA, RL reduces Prompt Robustness from 72.6 to 66.7, a 5.9-point drop that places it below the base model's score of 72.8, indicating severe generalization degradation after optimization in a fixed environment. In contrast, HSA-RL increases Prompt Robustness from 77.1 to 77.6 while also improving IFE-P and IFE-I. Overall, RL consistently strengthens task-specific performance, while HSA is critical for preventing overfitting to a fixed Harness and preserving generalization during RL.

\ifdefined\TRInlineSupplementarySections
\subsection{Training Dynamics}
\label{sec:training_dynamics}

\begin{figure}[!htbp]
\centering
\includegraphics[width=0.92\textwidth]{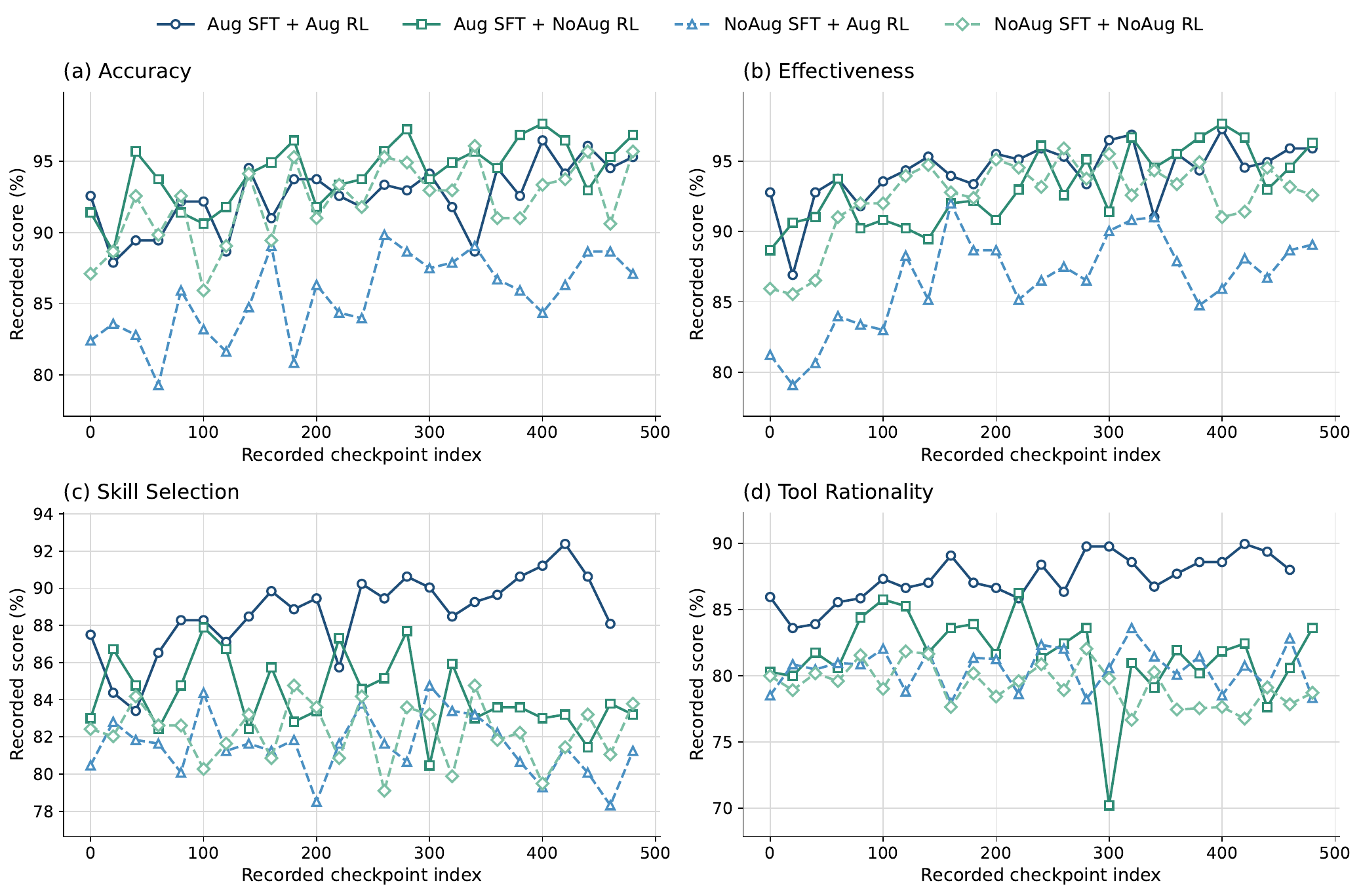}
\caption{Checkpoint trajectories across four training configurations formed by crossing HSA at the SFT and RL stages, evaluated on Accuracy, Effectiveness, Skill Selection, and Tool Rationality. In these trajectories, HSA-SFT is associated with higher reply-quality rewards, while HSA-RL is associated with stronger agentic-behavior rewards, especially Skill Selection and Tool Rationality; combining HSA-SFT and HSA-RL gives the most balanced late-stage reward profile.}
\label{fig:agentic_training_dynamics}
\end{figure}

Figure~\ref{fig:agentic_training_dynamics} shows that HSA-SFT is primarily associated with stronger reply-quality rewards, whereas HSA-RL raises the agentic-behavior dimensions, especially Skill Selection and Tool Rationality. Their combination gives the most balanced late-stage profile. These single-run trajectories are process diagnostics rather than evidence of training efficiency or cross-seed stability.

Figure~\ref{fig:cot_length_compression} traces the auxiliary CoT-length term in the selected HSA-RL run. From the first to the last 100-step window, mean extracted length decreases from 290.6 to 171.2 tokens for tool-call CoT and from 93.6 to 37.0 for final-reply CoT. The mean logged penalty score $q(L)$ moves from 0.466 to 0.272, corresponding to a raw reward contribution from $-0.0466$ to $-0.0272$ before GDPO normalization. The reduction is strongest early and then reaches a noisy plateau; it diagnoses trace length only, not held-out quality or causal latency gains.

\begin{figure}[!htbp]
\centering
\includegraphics[width=0.74\linewidth]{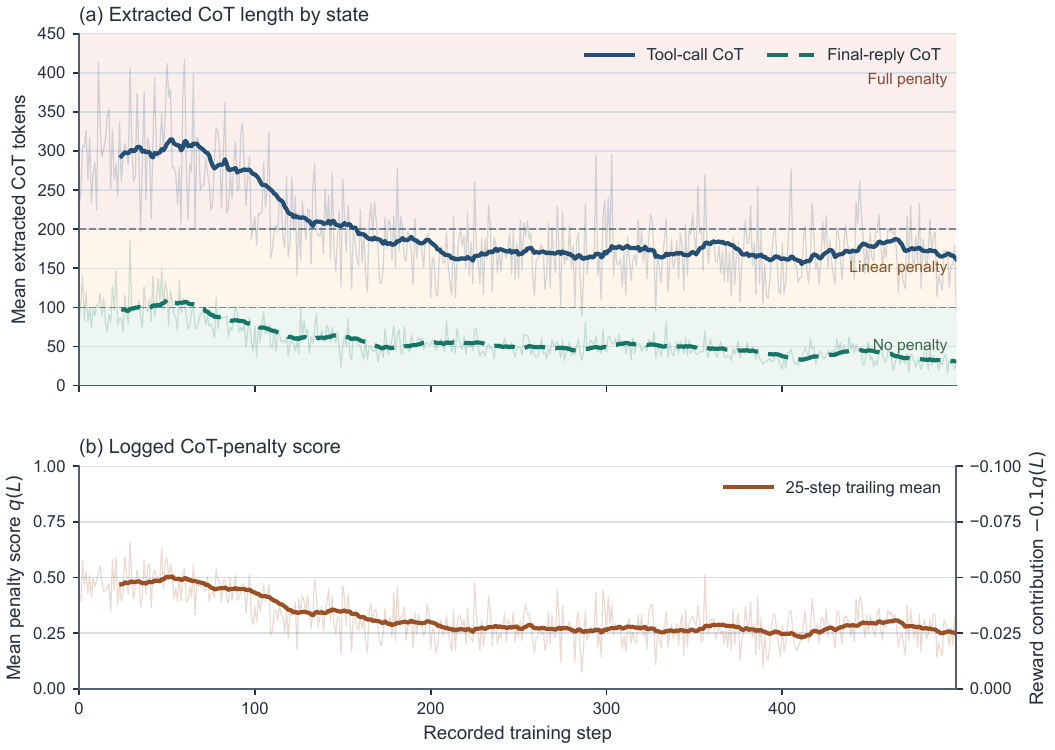}
\caption{Single-run CoT-length diagnostics for the selected HSA-RL trajectory. Faint lines show recorded step values and bold lines show 25-step trailing means. Panel (a) separates tool-call and final-reply CoT lengths and marks the no-penalty ($L\leq100$), linear ($100<L<200$), and saturated ($L\geq200$) regions. Panel (b) shows the logged score $q(L)\in[0,1]$ and the equivalent raw reward contribution $-0.1q(L)$. The figure is a training diagnostic, not a causal ablation of latency or quality.}
\label{fig:cot_length_compression}
\end{figure}

% Keep the diagnostic figure ordered, but allow the following deployment
% discussion to use any remaining space while the figure floats forward.

\fi

%% ============================================================
\subsection{Deployment Inference Performance}
\label{sec:latency}

Table~\ref{tab:latency} reports deployment inference performance in the application's low-concurrency operating range. Every row uses the same fixed 100-case sample and complete Agent workflow; the API rows traverse vendor-managed serving routes, while all H20 rows use one local GPU.

At concurrency 1 and 2, the selected MTP configuration reaches P95 latencies of 8.114 and 9.047\,s, and all measured requests satisfy the 15-second bound. On the same checkpoint and H20, MTP raises decoding throughput by 1.69$\times$ and 1.52$\times$, respectively. Cross-checkpoint Base--HAT Decode values are not intrinsic speed comparisons: each policy induces different Agent trajectories and completion-length distributions, and the arithmetic mean of per-call ratios is especially sensitive to very short generations. We therefore use the within-checkpoint Off--On contrast as the primary acceleration evidence. Under MTP On, HAT also has higher draft acceptance, so its remaining cross-checkpoint gap is not attributed to length alone. The offline quality values elsewhere in this section use MTP Off. The analyses below report MTP adaptation, same-checkpoint quality, and concurrency stress tests; Appendix~\ref{app:deployment_inference} retains workload, serving-configuration, and engine diagnostics. The API routes have P95 values above 21\,s at concurrency 1, but their hardware, batching, and service load are not observable; we use them only as end-to-end operational references.

\begin{table*}[t]
\captionsetup{type=table,hypcap=false}
\caption{Controlled low-concurrency deployment replay. Latencies are seconds; Decode is the arithmetic mean of client-observed completion tokens/s over model calls. It is length-sensitive and is not an intrinsic cross-checkpoint kernel-speed measure. TTFT is the first call's client-observed P95 and includes network and queueing. Each row contains 100 measured Agent requests after 10 warm-up cases. API rows are operational references, not same-hardware model comparisons.}
\label{tab:latency}
\footnotesize
\setlength{\tabcolsep}{4.0pt}
\begin{tabularx}{\textwidth}{@{}l *{7}{>{\centering\arraybackslash}X}@{}}
\toprule
\tablehead{Configuration} & \tablehead{MTP} & \tablehead{$C$} & \tablehead{Wall P50  (s)} & \tablehead{Wall P95 (s)} & \tablehead{TTFT P95 (s)} & \tablehead{Decode (tokens/s)} & \tablehead{$\leq$15s}  \\
\midrule
\rowcolor{TableGroup}\multicolumn{8}{@{}l}{\sffamily\bfseries\color{TableAccent}Vendor API} \\
DeepSeek-V4-flash & -- & 1 & 11.210 & 21.191 & 2.067 & 103.44 & 71\% \\
DeepSeek-V4-flash & -- & 2 & 11.976 & 26.278 & 1.291 & 97.01 & 71\% \\
DeepSeek-V4-pro & -- & 1 & 14.312 & 28.882 & 1.312 & 53.53 & 61\% \\
DeepSeek-V4-pro & -- & 2 & 13.752 & 23.393 & 1.248 & 55.04 & 60\% \\
\midrule
\rowcolor{TableGroup}\multicolumn{8}{@{}l}{\sffamily\bfseries\color{TableAccent}Qwen3.6-35B-A3B} \\
Base & Off & 1 & 4.101 & 10.172 & 0.508 & 195.42 & 100\% \\
Base & Off & 2 & 4.560 & 11.504 & 0.623 & 138.81 & 97\% \\
Base & On & 1 & 3.648 & 9.805 & 0.502 & 215.95 & 99\% \\
Base & On & 2 & 4.806 & 10.251 & 0.698 & 165.74 & 97\% \\
\midrule
\rowcolor{TableGroup}\multicolumn{8}{@{}l}{\sffamily\bfseries\color{TableAccent}HAT trained Qwen3.6-35B-A3B (Ours)} \\
Ours & Off & 1 & 4.176 & 8.98 & 0.499 & 160.30 & 100\% \\
Ours & Off & 2 & 4.936 & 10.074 & 0.692 & 129.22 & 100\% \\
\rowcolor{TableHighlight}\textbf{Ours} & \textbf{On} & \textbf{1} & \textbf{3.407} & \textbf{8.114} & 0.553 & \textbf{271.40} & \textbf{100\%} \\
\rowcolor{TableHighlight}\textbf{Ours} & \textbf{On} & \textbf{2} & \textbf{4.479} & \textbf{9.047} & 0.754 & \textbf{196.15} & \textbf{100\%} \\
\bottomrule
\end{tabularx}
\end{table*}

\ifdefined\TRInlineSupplementarySections
% \FloatBarrier
\subsubsection{MTP Draft-Head Adaptation}

We compare two standalone MTP-adaptation trajectories using sequences collected from HSA-RL policy rollouts. For the task-trained trajectory, the policy checkpoint starts from $S_0=$ HSA-SFT + General OPD, while the NextN MTP parameters are transplanted from Qwen3.6-35B-A3B. For the base trajectory, the policy and its native, factory-co-trained NextN head start from the default Qwen3.6-35B-A3B checkpoint. In both cases, the policy produces rollout labels and the MTP update is performed separately: policy parameters are treated as fixed by the MTP objective, and only the draft-head parameters are optimized. Both configurations use the same EAGLE-style speculative execution path at inference, so the comparison concerns draft-weight provenance and adaptation, not different speculative algorithms.

Let a rollout be $y_{1:T}$ and let $K$ be the number of future-token offsets. The MTP loss is the standard multi-token cross-entropy that predicts token $t+1+k$ from the prefix through $t$:
\begin{equation}
\begin{split}
\mathcal{L}_{\mathrm{MTP}}(\phi)
&= -\frac{1}{Z}\sum_{k=0}^{K-1}\sum_{t=1}^{T-1-k}
\log p_{\phi}^{(k)}\!\left(y_{t+1+k}\mid y_{\leq t}\right), \\
Z &= \sum_{k=0}^{K-1}(T-1-k),
\end{split}
\label{eq:mtp_loss}
\end{equation}
where $\phi$ denotes the MTP parameters and $k=0$ is next-token prediction. Thus RL supplies on-policy rollout sequences and labels, while the standalone MTP step uses only supervised multi-token CE; no reward or advantage enters Equation~\ref{eq:mtp_loss}.

\begin{figure}[!htbp]
\centering
\includegraphics[width=0.98\textwidth]{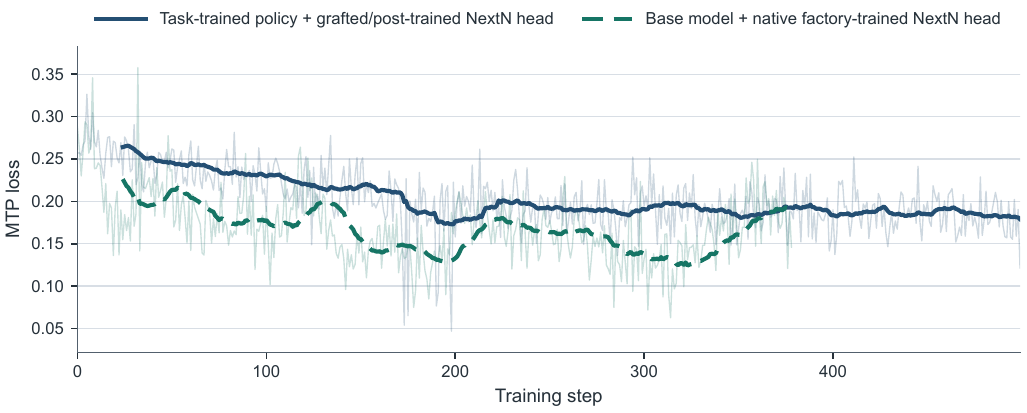}
\caption{Single-run standalone MTP-loss trajectories for a task-trained policy with a transplanted and subsequently post-trained NextN head, and for the base model with its native factory-trained head. Faint lines show step-level loss; bold lines show a 25-step trailing mean. The runs have different lengths and provide adaptation diagnostics only: they do not compare convergence speed, training stability, sample efficiency, or compute efficiency.}
\label{fig:mtp_loss_adaptation}
\end{figure}

Despite noisy per-step losses, the grafted-head trajectory moves from a higher early-loss region toward the late-training range of the native-head trajectory. This supports the practical conclusion that a task-trained policy can accept a base-model NextN head and adapt it through subsequent standalone training on policy-rollout labels. In an unplotted pilot, omitting the base MTP initialization produced a very high initial loss. We also observed that updating the main policy while leaving the MTP head untrained reduced draft acceptance to zero, consistent with the stale draft weights no longer predicting the updated policy. These are run-specific engineering observations rather than claims of universal or multi-seed convergence.

\subsubsection{MTP Quality and Concurrency Analysis}

Table~\ref{tab:mtp_concurrency} separates the deployment operating range (C=1--2) from stress conditions (C=4--8). At C=1--2, MTP increases the selected checkpoint's Decode TPS by 1.69$\times$/1.52$\times$ while retaining 100\% 15-second attainment. At C=4--8, the speedup falls to 1.27$\times$/1.19$\times$ and P95 no longer improves. The base model shows the same qualitative saturation boundary with a smaller low-concurrency gain. We therefore treat MTP as a low-concurrency latency optimization, not as a universal capacity multiplier.

\begin{table}[!htbp]
\centering
\caption{Same-checkpoint $\mathcal{T}_1$ quality check with MTP Off and On ($n=978$). Both rows use the same Final Evaluation Judge. AVG is the arithmetic mean of Accuracy and Effectiveness.}
\label{tab:mtp_quality_check}
\small
\setlength{\tabcolsep}{12pt}
\begin{tabular}{lccc}
\toprule
\tablehead{Serving mode} & \tablehead{Accuracy} & \tablehead{Effectiveness} & \tablehead{AVG} \\
\midrule
MTP Off & 95.8 & 93.8 & 94.8 \\
\rowcolor{TableHighlight}\textbf{MTP On} & \textbf{96.2} & \textbf{94.2} & \textbf{95.2} \\
\bottomrule
\end{tabular}
\end{table}

\begin{table}[!htbp]
\caption{MTP trade-off across concurrency. Arrows show Off $\rightarrow$ On. Decode TPS is client-observed mean per call; Wall and TTFT are P95 seconds. C=1, 2, and 8 are single runs. C=4 values are means of three run-level statistics, not pooled percentiles.}
\label{tab:mtp_concurrency}
\footnotesize
\setlength{\tabcolsep}{4.3pt}
\begin{tabular*}{\textwidth}{@{\extracolsep{\fill}}l c ccccc@{}}
\toprule
\tablehead{Checkpoint} & \tablehead{$C$} & \tablehead{Decode TPS} & \tablehead{Speedup} & \tablehead{Wall P95 (s)} & \tablehead{TTFT P95 (s)} & \tablehead{$\leq$15s} \\
\midrule
\multirow{4}{*}{Qwen3.6-35B-A3B}
 & 1 & 195.42$\rightarrow$215.95 & 1.11$\times$ & 10.172$\rightarrow$9.805 & 0.508$\rightarrow$0.502 & 100$\rightarrow$99\% \\
 & 2 & 138.81$\rightarrow$165.74 & 1.19$\times$ & 11.504$\rightarrow$10.251 & 0.623$\rightarrow$0.698 & 97$\rightarrow$97\% \\
 & 4 & 105.04$\rightarrow$113.25 & 1.08$\times$ & 13.408$\rightarrow$16.266 & 0.643$\rightarrow$1.088 & 97.3$\rightarrow$92.3\% \\
 & 8 & 70.28$\rightarrow$70.19 & 1.00$\times$ & 17.848$\rightarrow$26.136 & 0.953$\rightarrow$1.519 & 90$\rightarrow$79\% \\
\midrule
\multirow{4}{*}{HAT (Ours)}
 & 1 & 160.30$\rightarrow$271.40 & 1.69$\times$ & 8.98$\rightarrow$8.114 & 0.499$\rightarrow$0.553 & 100$\rightarrow$100\% \\
 & 2 & 129.22$\rightarrow$196.15 & 1.52$\times$ & 10.074$\rightarrow$9.047 & 0.692$\rightarrow$0.754 & 100$\rightarrow$100\% \\
 & 4 & 99.65$\rightarrow$126.64 & 1.27$\times$ & 10.869$\rightarrow$12.460 & 0.692$\rightarrow$1.349 & 99.3$\rightarrow$98.0\% \\
 & 8 & 59.01$\rightarrow$70.22 & 1.19$\times$ & 18.342$\rightarrow$20.298 & 0.900$\rightarrow$2.574 & 91$\rightarrow$75\% \\
\bottomrule
\end{tabular*}
\vspace{3pt}
\parbox{\textwidth}{\scriptsize\textit{C=4 run ranges (Off; On).} Qwen3.6 Base---Decode: 102.13--107.33; 111.11--116.49. Wall P95: 12.931--13.756; 14.957--18.054. TTFT P95: 0.563--0.758; 0.942--1.226. Attainment: 96--98\%; 89--95\%. Ours---Decode: 97.40--102.55; 121.61--130.18. Wall P95: 10.269--11.606; 10.532--14.205. TTFT P95: 0.587--0.784; 1.305--1.413. Attainment: 99--100\%; 96--100\%.}
\end{table}

\fi

\subsection{Held-Out Harness Edits Evaluation}
\label{app:harness_edit_benchmark}

We construct an additional held-out evaluation set for Harness edits by sampling 1{,}500 cases from real online live-streaming traffic, with 500 development cases and 1{,}000 held-out test cases. The raw logs contain viewer comments, dialogue history, live-room context, product information, tool-use traces, and model responses. We use an automatic quality check to count bad cases where the model output is not sufficiently natural or colloquial for a live-streaming host.

The development split is used only for Harness diagnosis and editing. We analyze the bad cases on the 500-case development set and revise editable Harness components, primarily the system prompt and Skills, while keeping the model weights fixed. The resulting Harness edits are fixed and applied unchanged to both the Fixed-Harness SFT checkpoint and the HAT checkpoint. The test split is held out from this editing loop and is used to compare each model before and after applying the same Harness edits.

On the held-out test set, the Harness edits reduce the total detected error count by 18.1\% for the Fixed-Harness SFT checkpoint and by 51.7\% for the HAT checkpoint, relative to their respective before-edit counts. Since lower error counts are better, these results show that the same Harness edits improve both models, with a substantially larger reduction for HAT.
\ifdefined\TRInlineSupplementarySections
Figure~\ref{fig:prompt_edit_case_analysis} in Appendix~\ref{app:harness_evolution} provides two qualitative examples in which the Fixed-Harness SFT checkpoint continues to expose internal uncertainty after the shared prompt edit, whereas HAT follows the revised Harness instruction. These cases illustrate edit compliance but do not constitute additional quantitative evidence.
\fi

% Downgrade shared files: \section->\subsection, \subsection->\subsubsection, \subsubsection->\paragraph
\begingroup
\let\TRorigsection\section
\let\TRorigsubsection\subsection
\let\TRorigsubsubsection\subsubsection
\let\section\TRorigsubsection
\let\subsection\TRorigsubsubsection
\let\subsubsection\paragraph
\section{Human Blind-Test Protocol and Results}
\label{app:blind_test}

We conduct a human blind test to compare the modular Harness runtime with a ReAct-style baseline on real live-streaming requests. We draw a stratified 100-request blind-test set with a fixed random seed so that its scenario composition follows the full pool.

For each request, both systems run independently on the same input. The record preserves the user input, dialogue history, complete reasoning and tool-use trace, and final response from both systems. During annotation, the two responses are randomly assigned to Model A and Model B, and the annotator is not shown the system identity. The annotator chooses one of three labels: Harness better, ReAct better, or tie. Non-tie decisions require a short reason. After annotation, all 100 examples are reviewed for data consistency and labeling consistency.

\begin{table}[t]
\centering
\begin{minipage}{0.78\linewidth}
\captionsetup{width=\linewidth}
\caption{Human blind-test preference distribution on 100 paired real live-streaming requests.}
\label{tab:blind_test_preference}
\small
\setlength{\tabcolsep}{5pt}
\begin{tabularx}{\linewidth}{@{}>{\raggedright\arraybackslash}X >{\centering\arraybackslash}p{0.22\linewidth} >{\centering\arraybackslash}p{0.22\linewidth}@{}}
\toprule
\tablehead{Preference} & \tablehead{Count} & \tablehead{Share} \\
\midrule
Harness better & 35 & 35.0\% \\
Tie & 64 & 64.0\% \\
ReAct better & 1 & 1.0\% \\
\bottomrule
\end{tabularx}
\end{minipage}
\end{table}

Table~\ref{tab:blind_test_preference} reports the preference distribution. Among the 36 examples with a clear preference, Harness wins 35, corresponding to a 97.2\% non-tie win rate. A two-sided binomial test against equal preference gives $p<0.001$. The only ReAct-preferred example is an ambiguous short comment, where Harness matches an irrelevant FAQ-style response while ReAct gives a safer transitional reply. We therefore treat this case as an isolated fallback-policy failure rather than a recurring disadvantage.

\begin{table}[t]
\centering
\begin{minipage}{0.78\linewidth}
\captionsetup{width=\linewidth}
\caption{Attribution of the 35 Harness-preferred examples. Categories are assigned from annotator rationales after the blind decision.}
\label{tab:blind_test_attribution}
\small
\setlength{\tabcolsep}{5pt}
\begin{tabularx}{\linewidth}{@{}>{\raggedright\arraybackslash}X >{\centering\arraybackslash}p{0.22\linewidth} >{\centering\arraybackslash}p{0.22\linewidth}@{}}
\toprule
\tablehead{Attribution category} & \tablehead{Count} & \tablehead{Share} \\
\midrule
More accurate input understanding & 12 & 34.3\% \\
More reliable output & 8 & 22.9\% \\
More appropriate scenario behavior & 8 & 22.9\% \\
Higher response quality & 5 & 14.3\% \\
More appropriate tool use & 2 & 5.7\% \\
\bottomrule
\end{tabularx}
\end{minipage}
\end{table}

Table~\ref{tab:blind_test_attribution} summarizes the attribution of Harness wins. The largest sources of improvement are input understanding, output reliability, and scenario-appropriate behavior, which together account for 80.0\% of the Harness-preferred cases. These categories include correctly separating live-room events from viewer questions, avoiding unsupported claims or false promises, handling low-information comments without forced product promotion, and using available tool evidence more appropriately. The blind-test result is consistent with the automatic Judge evaluation and provides an additional human preference check on real live-streaming requests.

\FloatBarrier
% 5.7--5.8 Deployment and Online A/B Test
\section{Online A/B Test on Taobao Live}
\label{sec:online_ab_taobao}

We deployed the Harness-based system in Taobao Live's production digital-avatar business and evaluated it against the existing ReAct system through a user-level online A/B test. Unlike the controlled replay above, this experiment measures real production traffic. The platform uses stable user-level bucketing with an 80/20 traffic allocation: the ReAct control contains 1{,}581{,}494 participating unique visitors (UVs), and the Harness treatment contains 395{,}276, for a total of 1{,}976{,}770. The two arms compare complete production system versions, so the contrast captures their joint effect rather than isolating the policy model, Harness, or routing configuration.

\needspace{4\baselineskip}
We assess product-detail-page engagement using Item Page View (IPV), the number of product-detail-page views. Because the two groups receive unequal traffic, their raw view counts are not directly comparable. The platform therefore normalizes IPV by the participating UV count. The relative uplift is $(V_{\mathrm{treat}}/N_{\mathrm{treat}})/(V_{\mathrm{ctrl}}/N_{\mathrm{ctrl}})-1$, where $V$ is the total IPV and $N$ is the number of participating UVs in each group.

\begin{table}[H]
\centering
\begin{minipage}{0.88\linewidth}
\captionsetup{width=\linewidth}
\caption{UV-normalized item-page-view result from the online A/B test in Taobao Live's production digital-avatar business. The relative uplift compares the Harness treatment with the ReAct control; the significance assessment is reproduced from the experiment platform.}
\label{tab:taobao_online_ab}
\small
\setlength{\tabcolsep}{5pt}
\begin{tabularx}{\linewidth}{@{}>{\raggedright\arraybackslash}X c c@{}}
\toprule
\tablehead{Metric per participating UV} & \tablehead{Relative uplift} & \tablehead{Platform assessment} \\
\midrule
Item Page View (IPV) & +0.9107\% & Significantly positive \\
\bottomrule
\end{tabularx}
\vspace{2pt}

\parbox{\linewidth}{\scriptsize\textit{Note:} The dashboard records one positively significant day for IPV. The supplied snapshot does not expose the statistical test, threshold, $p$-values, or confidence intervals; we therefore reproduce the platform's assessment without inferring a specific significance level or a long-term effect.}
\end{minipage}
\end{table}

As shown in Table~\ref{tab:taobao_online_ab}, the Harness treatment increases IPV per participating UV by 0.9107\% relative to ReAct, an effect classified as significantly positive by the experiment platform. This result indicates improved product-detail-page engagement for the deployed Harness system as a whole during the observed test window.

\endgroup

% \FloatBarrier
% 6. Related Work
% \input{sections/related_work}

\FloatBarrier
% 7. Conclusion
\section{Conclusion}
\label{sec:conclusion}

We presented Harness-Aware Training (HAT), a systems-and-training methodology for the tension between versioned Harness change and policy stability in live-streaming digital-avatar agents. It treats Harness-state variation as a training-distribution design problem within an explicitly defined change envelope.

Our work makes four contributions. First, Harness Evolution versions runtime changes separately from policy updates through modular Skills, Hooks, prompts, and tools, and exposes the training problem created by a moving execution environment. Second, HAT treats Harness-state variation as part of the training distribution through HSA and a three-stage pipeline with HSA-SFT, General OPD, and HSA-RL. Third, scoped offline sets, a human-aligned evaluation panel, and a controlled complete-Agent deployment replay separate checkpoint quality evidence from serving-route behavior. Fourth, the ablations provide practical training insights: General OPD mainly recovers instruction-following ability after Fixed-Harness SFT, while HSA is most effective when introduced during SFT.

In our study, the HAT-trained compact model reaches 94.8 Live-Stream QA AVG and does not exhibit the general-set drop observed after Fixed-Harness SFT. On one H20 with MTP enabled, the controlled replay reaches 8.114\,s P95 at concurrency 1 and 9.047\,s at concurrency 2, with 100\% 15-second attainment; MTP increases client-observed decoding throughput by 1.69$\times$/1.52$\times$ over the same checkpoint without MTP. An online A/B test in Taobao Live's production digital-avatar service further shows an increase in item-page views per participating user relative to ReAct. Overall, these results show that HAT can produce a latency-feasible compact agent while preserving the adaptability needed for evolving production Harnesses.

\FloatBarrier
% Authors
\section*{Author Contributions}
\noindent\textbf{Project Leader:} Yuhan Sun\\
\textbf{Core Contributors:} Wenhao Lin, Yongdong Luo, Yibo Hu, Junfeng Ma, Meiguang Jin\\
\textbf{Contributors:} Weihang Pan, Jiaxin Zhao, Zulong Chen\\[0.3em]

\FloatBarrier
% Acknowledgments
\section*{Acknowledgments}
We thank the Alibaba ROLL Team for their invaluable support on training infrastructure. We are grateful to Dr.\ Weihang Pan from Zhejiang University and Zulong Chen from Alibaba Group for their highly valuable contributions and assistance.

\FloatBarrier
% References
\bibliographystyle{plainnat}
\bibliography{references}

\FloatBarrier
% Appendix
\appendix
\setcounter{topnumber}{4}
\setcounter{bottomnumber}{2}
\setcounter{totalnumber}{6}
\makeatletter
\setlength{\@fpsep}{10pt plus 2pt minus 2pt}
\setlength{\@fpbot}{0pt plus 1fil}
\makeatother
\section{Runtime Interfaces and Skills}
\label{app:runtime_inventories}
\newenvironment{TRruntimeTable}{\begin{table}[!htbp]}{\end{table}}
\ifdefined\TRruntimeTable\else
\newenvironment{TRruntimeTable}{\begin{table*}[t]}{\end{table*}}
\fi
\begin{TRruntimeTable}
\centering
\caption{Working inventory of built-in and MCP-discovered runtime interfaces.}
\label{tab:runtime_interfaces}
\footnotesize
\setlength{\tabcolsep}{4pt}
\begin{tabularx}{\textwidth}{@{}p{2.35cm} >{\raggedright\arraybackslash}X >{\raggedright\arraybackslash}p{5.1cm}@{}}
\toprule
\tablehead{Responsibility} & \tablehead{Interfaces} & \tablehead{Role} \\
\midrule
Product retrieval &
\path{search_product_by_keyword}, \path{get_current_product_info}, \path{get_product_info_by_link_id}, \path{get_product_extra_info_by_link_id_and_keywords}, \path{search_preset_faq} &
Find the active or referenced product and retrieve catalog, detail-page, knowledge-base, or FAQ evidence. \\
Pricing and promotion &
\path{get_price_info_by_link_id}, \path{get_promotion_info} &
Retrieve SKU-level prices, entitlements, coupons, and room-level promotions. \\
Action and flow control &
\path{change_explain_order}, \path{load_skill}, \path{get_current_time}, \path{refuse_to_reply} &
Change presentation order, load behavioral instructions, resolve time-sensitive rules, or silently discard meaningless input. \\
MCP lookup &
\path{query_promotion}, \path{query_buyer_resource}, \path{query_fund_asset}, \path{query_coupon_detail} &
Retrieve campaign conditions and buyer-side resources from the external marketing service. \\
MCP calculation &
\path{calculate_optimal_promotion}, \path{calculate_promotion} &
Compute eligible promotional combinations and resulting prices. \\
\bottomrule
\end{tabularx}
\end{TRruntimeTable}

\ifdefined\TRruntimeTable\else
\newenvironment{TRruntimeTable}{\begin{table*}[t]}{\end{table*}}
\fi
\begin{TRruntimeTable}
\centering
\caption{Working Skill inventory and routing responsibilities.}
\label{tab:skill_inventory}
\footnotesize
\setlength{\tabcolsep}{4pt}
\begin{tabularx}{\textwidth}{@{}>{\raggedright\arraybackslash}p{2.6cm} >{\raggedright\arraybackslash}X >{\raggedright\arraybackslash}p{3.25cm} >{\raggedright\arraybackslash}X@{}}
\toprule
\tablehead{Reply Skill} & \tablehead{Response responsibility} & \tablehead{Strategy Skill} & \tablehead{Tool-chain responsibility} \\
\midrule
\path{item_qa} & Product attributes, specifications, comparison, price, and recommendation. & \path{tool_strategy_benefit} & Entitlement, price, and promotion retrieval. \\
\path{chillchat} & Non-product conversation and engagement. & \path{tool_strategy_compare} & Multi-product comparison and retrieval. \\
\path{aftersale} & Returns, exchanges, complaints, and other after-sales requests. & \path{tool_strategy_multi} & Aggregation and orchestration for multiple comments. \\
\path{clarification} & Follow-up questions when available context is insufficient. & \path{tool_strategy_switch} & Presentation-order adjustment. \\
\path{faq_reply} & Responses grounded in configured FAQ entries. & \path{general_discount} & Room-level promotion retrieval and response composition. \\
\path{refusal} & Declines for inappropriate or unsupported requests. & & \\
\path{change_order} & Confirmation of presentation-order changes. & & \\
\path{thanks} & Responses to gratitude and positive feedback. & & \\
\path{greet} & Welcome messages for viewers entering the room. & & \\
\bottomrule
\end{tabularx}
\end{TRruntimeTable}

\section{Harness Evolution}
\label{app:harness_evolution}
\subsection{Evolution Protocol}

Each Harness Evolution stage holds model fixed and changes only editable runtime modules. The evolution loop contains five steps:
\begin{enumerate}[leftmargin=*,itemsep=2pt]
\item \textbf{AI diagnosis.} Analyze the latest evaluation result, cluster the top failure causes, attach representative bad cases, and propose root causes.
\item \textbf{Human confirmation.} A developer reviews the proposed skill, prompt, hook, or tool changes, then accepts, modifies, or rejects the plan.
\item \textbf{AI-assisted editing.} Apply the confirmed changes to skill definitions, prompt assembly, Hooks, and tool logic while preserving a versioned configuration snapshot.
\item \textbf{Human-triggered evaluation.} Run inference and scoring on the dev-set.
\item \textbf{Regression check and planning.} Compare paired category changes, inspect new regressions, and let a human choose promotion, rollback, another evolution stage, or stopping.
\end{enumerate}
The stopping rule is operational: stop when the remaining errors are sparse long-tail cases and local rule additions are more likely to cause cross-category regression than systematic improvement.

\subsection{Evolution-Stage Changes and Diagnostics}

\begin{table}[!htbp]
\caption{Fixed-policy Harness Evolution record. Values come from six direct evaluation summaries on the same 482-item dev-set. ``Selected'' is an engineering early-stop decision.}
\label{tab:harness_evolution_details}
\footnotesize
\setlength{\tabcolsep}{4pt}
\begin{tabularx}{\textwidth}{@{}l >{\hsize=1.25\hsize\raggedright\arraybackslash}X cc >{\hsize=.75\hsize\raggedright\arraybackslash}X@{}}
\toprule
\tablehead{Stage} & \tablehead{Harness changes} & \tablehead{Acc.} & \tablehead{Eff.} & \tablehead{Engineering diagnosis} \\
\midrule
ReAct & Conventional reasoning--action loop without the modular Harness. & 80.33 & 84.58 & System baseline. \\
Harness base & Initial manually maintained Skills and modular runtime. & 82.40 & 87.16 & Establishes the editable Harness. \\
Evolution 1 & Add refusal tool and stop-loop Hook; rewrite refusal, chat, and after-sales Skills; add four global constraints. & 92.13 & 84.16 & Accuracy rises sharply; internal diagnosis attributes Effectiveness loss to over-triggered refusal. \\
\rowcolor{TableHighlight}\textbf{Evolution 2} & Add seven whitelist exclusions before refusal; map explanation triggers and require the relevant tool call. & \best{92.55} & \best{92.75} & \textbf{Selected}; restores Effectiveness while retaining Accuracy. \\
Evolution 3 & Add attribution, factuality, parameter-completeness, tool-use, and system-message rules; expand refusal exclusions. & 91.51 & 90.89 & Long-tail rules interact and regress both metrics. \\
Evolution 4 & Relax length, tool-trigger, parameter, transaction-intent, and attribution restrictions. & 91.51 & 89.96 & Partial rollback does not recover Effectiveness. \\
\bottomrule
\end{tabularx}
\end{table}

The process report attributes the largest Accuracy gains to negative-feedback chat (+60.7 points), after-sales and returns (+53.3), and purchase/payment cases (+33.3). The largest Effectiveness gains occur for emotion expression (+27.3), membership or gift benefits (+18.8), and positive-feedback or already-purchased cases (+14.3). These category values are engineering diagnostics rather than pre-specified inferential analyses.

After Evolution~2, the report records 36 remaining Accuracy failures across 16 categories and 35 Effectiveness failures without a dominant systematic cluster. Evolutions~3--4 show why ``more rules'' is not equivalent to better evolution: targeted long-tail fixes create interactions across Skills, Hooks, and global instructions. This observation motivates configuration diversity during training and regression-aware early stopping during production evolution.

\Needspace{0.58\textheight}
\subsection{Prompt-Edit Case Analysis}

\begin{figure}[!htbp]
\centering
\includegraphics[width=\textwidth]{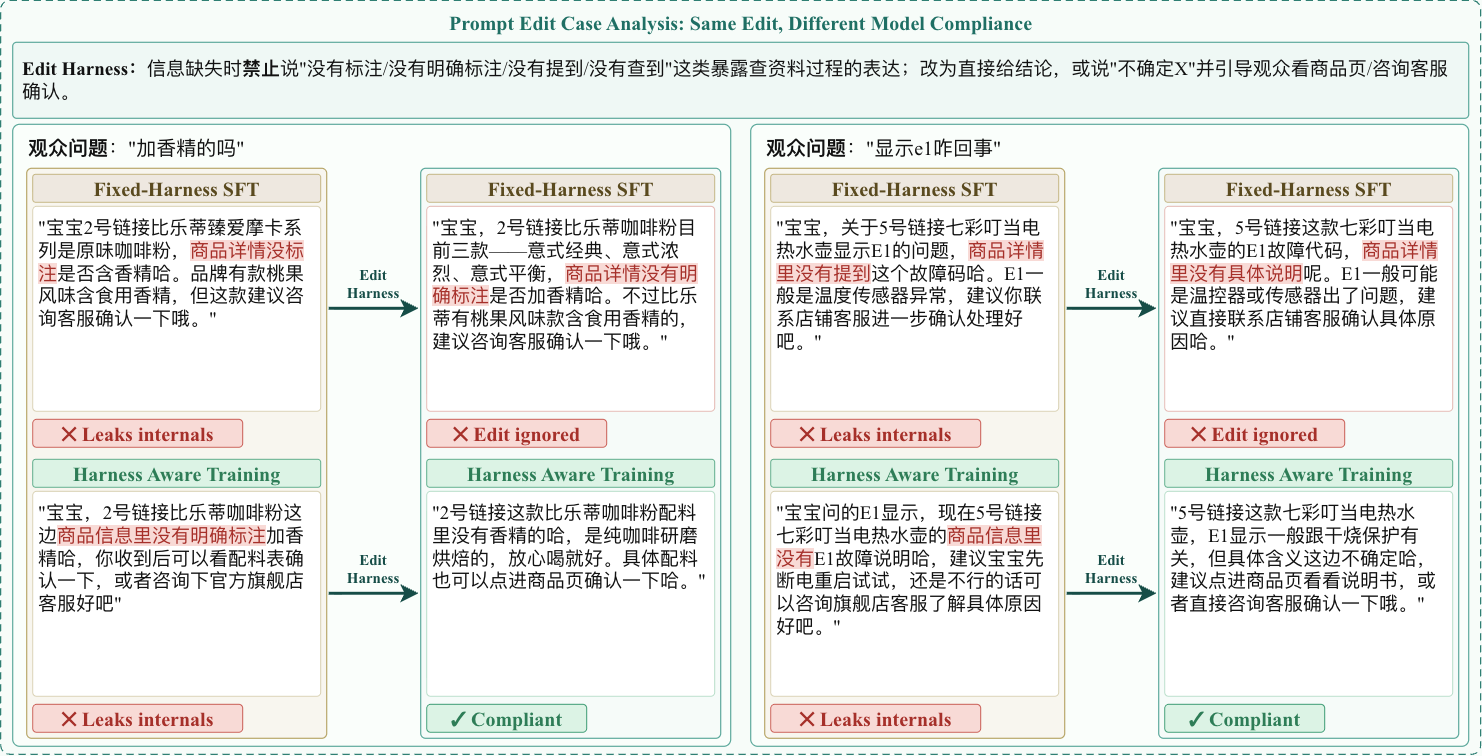}
\caption{Qualitative prompt-edit cases under the same Harness revision. The edit prohibits disclosure-process language when the available evidence does not support a definite conclusion. In both examples, the Fixed-Harness SFT checkpoint continues to expose internal uncertainty after the edit, whereas HAT follows the revised instruction and gives a direct response or routes the viewer to product details or customer service. Red text marks the disclosure phrases targeted by the edit. These examples illustrate compliance behavior and are not additional quantitative evidence.}
\label{fig:prompt_edit_case_analysis}
\end{figure}
\FloatBarrier

\subsection{Judge Evolution Case Study}
\label{app:judge_evolution}

\begin{figure}[!htbp]
\centering
\includegraphics[width=0.6\linewidth]{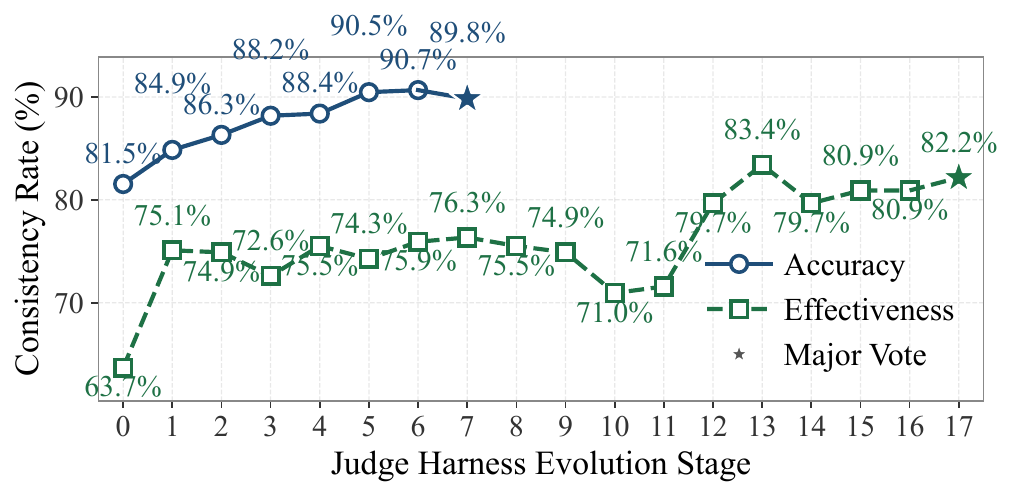}
\caption{Judge alignment to a human-labeled calibration cohort ($n=482$). Accuracy and Effectiveness agreement improve through evidence-tool, rubric, and voting revisions; the cohort evaluates the scoring instrument rather than the policy.}
\label{fig:judge_calibration}
\end{figure}

The Final Evaluation Judge is a Harness-based panel that aggregates DeepSeek-V4-pro, GLM-5.2, and Qwen3.7-max by majority vote~\citep{pan2026fgd}. Its Skills, prompts, and tool-verification procedures are evolved through disagreement analysis against human labels. Figure~\ref{fig:judge_calibration} shows the trajectory on 482 human-labeled interactions: Accuracy agreement moves from 81.54\% at Evolution~1 to 90.46\% at Evolution~6, while Effectiveness moves from a 63.70\% uncalibrated baseline to an 83.40\% peak. The selected multi-model voting configuration reaches 82.2\% Effectiveness agreement. Disagreement clusters include missed multi-comment intents, unsupported capability claims, incorrect product linkage, and inconsistent grading of partial answers; revisions target these recurring classes rather than individual model outputs.

\FloatBarrier
\section{Evaluation and Reproducibility}
\label{app:evaluation_details}
\subsection{Judge Roles and Independence}
\label{app:judge_roles}

Five roles must not be conflated: the DeepSeek-V4-pro SFT teacher generates supervised candidates; the rejection-sampling Judge filters them; the DeepSeek-V4-flash Reward Judge scores HSA-RL trajectories; the Final Evaluation Judge produces offline scores by majority vote over DeepSeek-V4-pro, GLM-5.2, and Qwen3.7-max; and human labels provide the calibration reference.

The Reward and Final Evaluation Judges have zero model overlap: DeepSeek-V4-flash is not a member of the final three-model panel. DeepSeek-V4-pro is both SFT teacher and one final voter, but a two-of-three majority prevents that vote from deciding any item alone. GLM-5.2 and Qwen3.7-max supply two non-teacher votes, and all three voters can call evidence tools when checking product facts. Human alignment is measured before the panel is used for model comparison.
%Section~\ref{sec:limitations} states the remaining boundary of calibration-set reuse and teacher-style preference.

\subsection{Offline Scoring Implementation}

Accuracy is binary; Effectiveness uses 0, 0.5, and 1; AVG is the per-sample mean of the two computed before rounding. $\mathcal{T}_3$ reports Tool and Prompt robustness separately, and IFEval uses its official prompt- and instruction-level evaluators. The policy is Qwen3.6-35B-A3B. HSA-RL uses GRPO with Accuracy, Effectiveness, Tool Rationality, and Skill Selection rewards under Skill, tool, prompt, Hook, and message-processing perturbations. All offline model comparisons use identical set versions and the same Final Evaluation Judge.

\subsection{\texorpdfstring{$\mathcal{T}_3$}{T3} Robustness Set Construction}
\label{app:t3_robustness}

$\mathcal{T}_3$ contains two synthetic robustness subsets generated from a real-business scenario seed library: Tool Robustness ($n=1{,}532$) and Prompt Robustness ($n=491$). The seed library is built from live-streaming e-commerce scenarios and records the product context, viewer intent, possible tool needs, and reply constraints for each scenario family. 

\paragraph{Construction pipeline} We use a shared seed-driven generation pipeline and instantiate it into two orthogonal branches. The seed pools are extracted from real live-streaming business data, including product context, FAQ-style knowledge, dialogue history, and historical tool-use traces. For each item, an LLM generator expands a sampled business seed into a natural viewer comment, live-room or product context, and the expected behaviors to be evaluated. The generated item is then converted into the Harness input format with the corresponding tool configuration, instruction configuration, simulated tool-return map, and evaluation rubrics. We apply rule-based and LLM-based consistency checks to ensure that the query, context, tool returns, instructions, and rubrics are mutually consistent. Items that fail validation are discarded and regenerated.

The Tool Robustness subset focuses on tool selection under distractor tools, while the Prompt Robustness subset additionally constructs instruction-following prompts to test whether the model can satisfy valid prompt constraints. The corresponding metrics are  Tool Called (TC), Response Contains (RC), and Addresses Core Query (ACQ) for Tool Robustness, and Addresses Core Query (ACQ), Follows Instructions (FI), and Tool Call Order (TCO) for Prompt Robustness. Each final item may contain multiple rubrics. AVG is the fraction of passed rubrics computed per item before aggregation. Overall, $\mathcal{T}_3$ tests model robustness to unseen in-domain tools and instructions in synthetic live-streaming scenarios.

\subsection{Bootstrap Stability Analysis}
\label{app:bootstrap_stability}

Table~\ref{tab:independent_bootstrap_ci} reports an independent paired-bootstrap check on a separate fixed evaluation run with cached model outputs. The intervals are computed over evaluation items using 10{,}000 paired bootstrap resamples. Because this run regenerates model answers rather than reusing the exact outputs underlying the main tables, small differences from the main-table point estimates arise from generation stochasticity. These results are therefore used as a stability check for the qualitative contrasts rather than as replacements for the reported main results. $P_{\mathrm{boot}}(\Delta \le 0)$ denotes the fraction of bootstrap replicates whose contrast is non-positive.

\begin{table}[!htbp]
\caption{Independent paired-bootstrap stability check. Differences are reported in percentage points; intervals are 95\% percentile bootstrap intervals over evaluation items.}
\label{tab:independent_bootstrap_ci}
\footnotesize
\setlength{\tabcolsep}{3.5pt}
\begin{tabularx}{\linewidth}{@{}>{\raggedright\arraybackslash}X >{\centering\arraybackslash}p{0.35\linewidth} >{\centering\arraybackslash}p{0.25\linewidth}@{}}
\toprule
\tablehead{Contrast / dimension} & \tablehead{$\Delta$ [95\% CI]} & \tablehead{$P_{\mathrm{boot}}(\Delta \le 0)$} \\
\midrule
\rowcolor{TableGroup}\multicolumn{3}{@{}l}{\sffamily\bfseries\color{TableAccent}IFEval prompt-level accuracy} \\
Fixed-Harness SFT $-$ Base & $-6.47\;[-10.72,\;-2.22]$ & $0.9992$ \\
Ours $-$ Base & $+2.03\;[-1.11,\;+5.18]$ & $0.1134$ \\
Ours $-$ Fixed-Harness SFT & $+8.50\;[+4.44,\;+12.38]$ & $0.0000$ \\
\midrule
\rowcolor{TableGroup}\multicolumn{3}{@{}l}{\sffamily\bfseries\color{TableAccent}Prompt Robustness} \\
Ours $-$ Fixed-Harness SFT & $+8.46\;[+5.27,\;+11.66]$ & $<0.0001$ \\
Fixed-Harness SFT $-$ Base & $-3.22\;[-6.59,\;+0.21]$ & $0.9661$ \\
\midrule
\rowcolor{TableGroup}\multicolumn{3}{@{}l}{\sffamily\bfseries\color{TableAccent}$\mathcal{T}_1$ Live-Stream QA: Ours $-$ Base} \\
Accuracy & $+6.95\;[+4.60,\;+9.41]$ & $<0.0001$ \\
Effectiveness & $+30.88\;[+27.76,\;+33.90]$ & $<0.0001$ \\
AVG & $+18.92\;[+17.00,\;+20.81]$ & $<0.0001$ \\
\bottomrule
\end{tabularx}
\end{table}

\FloatBarrier
\section{Deployment and Serving Details}
\label{app:deployment_inference}
\label{app:deployment_details}
\label{app:supplementary_details}

\subsection{Deployment Workload Characteristics}
\label{app:deployment_workload}
\begin{table}[!htbp]
\centering
\caption{C=1 workload shape. Values are means over successful complete-Agent requests or their constituent model calls.}
\label{tab:deployment_workload}
\footnotesize
\setlength{\tabcolsep}{4.5pt}
\begin{tabularx}{\linewidth}{@{}l *{4}{>{\centering\arraybackslash}X}@{}}
\toprule
\tablehead{Route / checkpoint} & \tablehead{MTP} & \tablehead{Calls/req} & \tablehead{Input/call} & \tablehead{Output/call} \\
\midrule
DeepSeek-V4-flash API & -- & 4.04 & 6,052 & 146 \\
DeepSeek-V4-pro API & -- & 3.67 & 5,537 & 119 \\
Qwen3.6-35B-A3B & Off & 3.23 & 5,622 & 115 \\
Qwen3.6-35B-A3B & On & 3.08 & 5,571 & 109 \\
HAT (Ours) & Off & 3.04 & 5,679 & 106 \\
\rowcolor{TableHighlight}\textbf{HAT (Ours)} & \textbf{On} & \textbf{3.12} & \textbf{5,786} & \textbf{105} \\
\bottomrule
\end{tabularx}
\end{table}

% \subsection{Resource-Accounting Contract}

% Each stage should be associated with an immutable run ID and report start/end time, hardware type and count, parallelism, cumulative GPU-hours, wall-clock hours, student input/output and optimizer-update tokens, teacher input/output tokens, environment interactions and tokens, external-model requests and monetary cost, invalid trajectories, infrastructure retries, and retry cost. HSA-SFT must include candidate generation and rejection filtering; General OPD must include base-teacher inference; HSA-RL must include the simulator, tools, Reward Judge, and failed rollouts. Final evaluation is reported separately from training. These fields prevent a nominally equal wall-clock budget from hiding different service-side or teacher costs.

\subsection{MTP Serving Configuration}

The MTP-On point estimates are not lower on either quality dimension, so this run shows no aggregate quality degradation from enabling the adapted draft head. Because sample-level paired outputs and repeated decoding runs are not available, Table~\ref{tab:mtp_quality_check} is not an equivalence test. Table~\ref{tab:mtp_serving_config} records the exact MTP serving configuration used for the reported measurements.

\begin{table}[!htbp]
\centering
\caption{MTP serving configuration used for the reported quality check and deployment measurements.}
\label{tab:mtp_serving_config}
\small
\setlength{\tabcolsep}{7pt}
\begin{tabularx}{0.92\textwidth}{@{}>{\raggedright\arraybackslash}p{0.31\textwidth} >{\ttfamily\raggedright\arraybackslash}X >{\raggedleft\arraybackslash}p{0.08\textwidth}@{}}
\toprule
\tablehead{Setting} & \normalfont\tablehead{Engine argument} & \tablehead{Value} \\
\midrule
Speculative algorithm & -{}-speculative-algo & NEXTN \\
Speculative steps & -{}-speculative-num-steps & 3 \\
EAGLE top-$k$ & -{}-speculative-eagle-topk & 1 \\
Draft tokens & -{}-speculative-num-draft-tokens & 4 \\
Single-token acceptance threshold & -{}-speculative-accept-threshold-single & 0.5 \\
Accumulated acceptance threshold & -{}-speculative-accept-threshold-acc & 0.7 \\
\bottomrule
\end{tabularx}
\end{table}

\FloatBarrier
\subsection{SGLang MTP Diagnostics}

\begin{table}[!htbp]
\caption{SGLang log-sample diagnostics for MTP-On serving. Triples are mean/P50/P95.}
\label{tab:sglang_mtp_diagnostics}
\footnotesize
\setlength{\tabcolsep}{2.8pt}
\begin{tabularx}{\linewidth}{@{}l *{3}{>{\centering\arraybackslash}X}@{}}
\toprule
\tablehead{Checkpoint} & \tablehead{Accept len} & \tablehead{Accept rate} \\
\midrule
Qwen3.6-35B-A3B & 2.94/2.92/3.58 & .735/.730/.890 \\
\rowcolor{TableHighlight}HAT (Ours) & 3.12/3.12/3.62 & .780/.780/.910 \\
\bottomrule
\end{tabularx}
\end{table}

As shown in Table~\ref{tab:sglang_mtp_diagnostics}.  From Base to HAT, mean accept length moves from 2.94 to 3.12 and mean acceptance rate from 0.735 to 0.780; these diagnostics provide an additional mechanism-level explanation for the MTP-On cross-checkpoint TPS difference beyond output length. Acceptance measures speculative-token verification, not semantic answer quality.

\end{document}